\documentclass[10pt,twocolumn,letterpaper]{article}

\usepackage[pagenumbers]{cvpr} 

\usepackage{graphicx}
\usepackage{amsmath}
\usepackage{amssymb}
\usepackage{booktabs}
\usepackage{bm}
\usepackage{tabularx}
\usepackage{array}
\usepackage{float}
\usepackage{wrapfig}
\usepackage{multirow}

\DeclareMathAlphabet{\mathdutchcal}{U}{dutchcal}{m}{n}
\SetMathAlphabet{\mathdutchcal}{bold}{U}{dutchcal}{b}{n}
\DeclareMathAlphabet{\mathdutchbcal}{U}{dutchcal}{b}{n}

\usepackage[pagebackref,breaklinks,colorlinks]{hyperref}

\usepackage[capitalize]{cleveref}
\crefname{section}{Sec.}{Secs.}
\Crefname{section}{Section}{Sections}
\Crefname{table}{Table}{Tables}
\crefname{table}{Tab.}{Tabs.}

\newcommand{\vt}[1]{\mathbf{#1}}
\newcommand{\mat}[1]{\bm{#1}}

\newcommand{\I}{\mathcal{I}}

\DeclareMathOperator*{\med}{med}

\usepackage{ulem}
\usepackage{afterpage}

\def\cvprPaperID{2143} 
\def\confName{CVPR}
\def\confYear{2024}

\begin{document}

\title{Erasing the Ephemeral: Joint Camera Refinement and Transient Object \\Removal
for Street View Synthesis}

\author{Mreenav Shyam Deka$^{*}$ \quad Lu Sang$^{*}$ \quad Daniel Cremers \\
Technical University of Munich\\
Computer Vision Group\\
{\tt\small mreenav.deka, lu.sang, cremers@tum.de}
}
\maketitle

\let\thefootnote\relax\footnote{$^{*}$ These authors contributed equally.}
\begin{abstract}
Synthesizing novel views for urban environments is crucial for tasks like autonomous driving and virtual tours. Compared to object-level or indoor situations, outdoor settings present unique challenges such as inconsistency across frames due to moving vehicles and camera pose drift over lengthy sequences. In this paper, we introduce a method that tackles these challenges on view synthesis for outdoor scenarios.
We employ a neural point light field scene representation and strategically detect and mask out dynamic objects to reconstruct novel scenes without artifacts. 
Moreover, we simultaneously optimize camera pose along with the view synthesis process, and thus we simultaneously refine both elements.
Through validation on real-world urban datasets, we demonstrate state-of-the-art results in synthesizing novel views of urban scenes.  
\end{abstract}

\section{Introduction}
\afterpage{
\begin{figure}[t!]
\setlength{\tabcolsep}{1pt}
\newcolumntype{Y}{>{\centering\arraybackslash}p{0.32\linewidth}}
    \centering
    \begin{tabular}{YYY}
        \includegraphics[width=\linewidth]{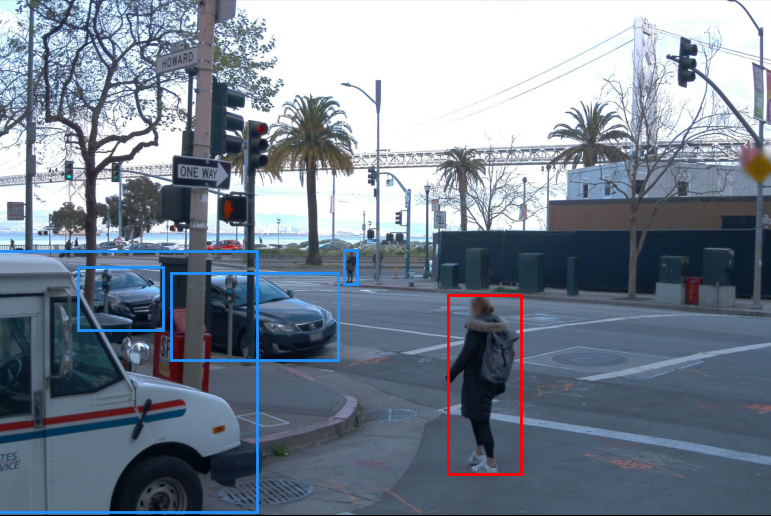} &
        \includegraphics[width=\linewidth]{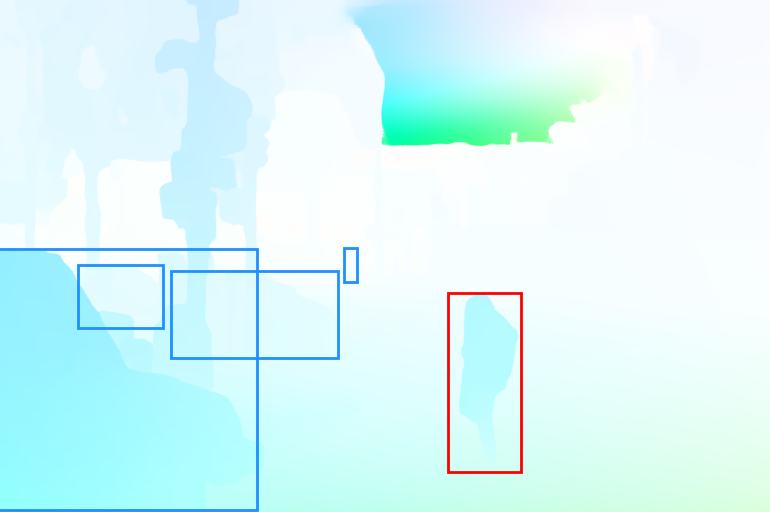} & 
        \includegraphics[width=\linewidth]{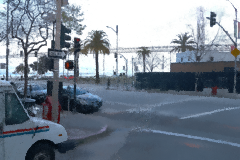}\\
        Detected Object & Optical Flow & \small Reconstructed scene \\
        \includegraphics[width=\linewidth]{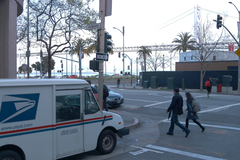} &
        \includegraphics[width=\linewidth]{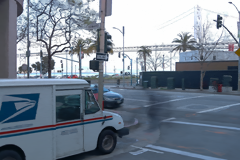} &
        \includegraphics[width=\linewidth]{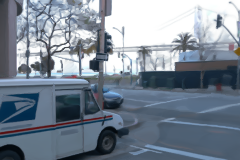} \\
         Unseen View & W/o erasing & Synthesized scene \\
         \includegraphics[width=\linewidth]{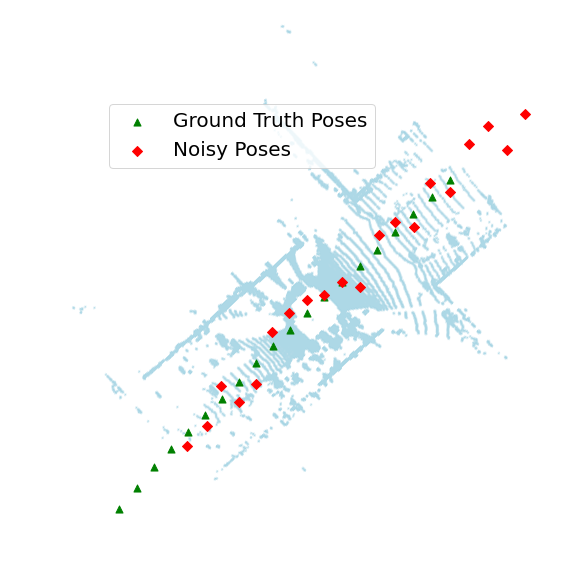} & 
         \includegraphics[width=\linewidth]{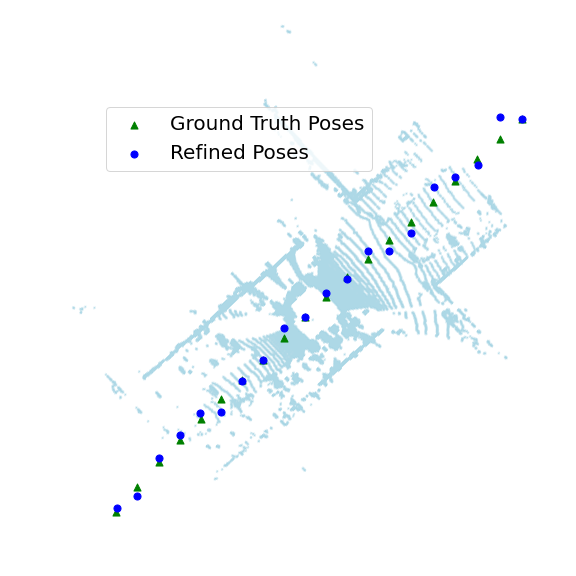} & 
         \includegraphics[width=\linewidth]{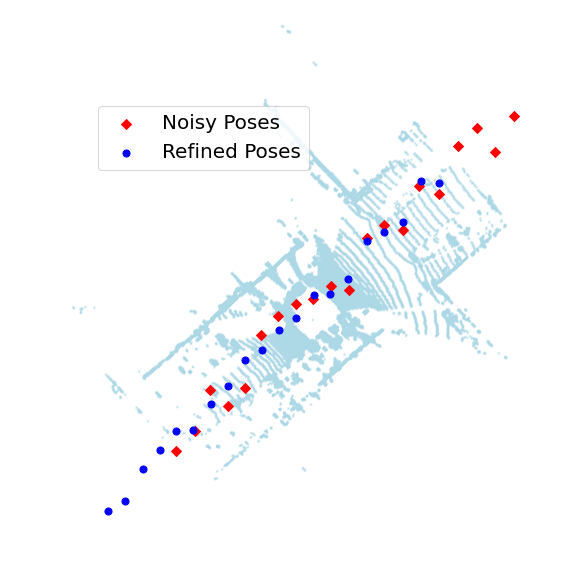}\\
         Noisy pose & Refined pose & Comparison 
    \end{tabular}
    \caption{The proposed method detects objects in frames and uses a voting scheme on the estimated optical flow to identify the transient object (first row). Our method erases moving objects in the rendered view and the artifacts created by them, leading to clearer renderings when compared to non-erasing methods (second row). The camera poses are refined jointly (third row).}
    \label{fig:teaser}
    \vspace{-0.5cm}
\end{figure}
}
Street view synthesis is an important part of digital mapping solutions that provides continuous views of streets. This has found numerous applications in areas such as virtual tours and navigation. A previous approach to virtually exploring streets involves overlapping and stitching together images captured from different view angles and tiling them at different zoom levels~\cite{streetview}.
However, this limits users to only viewing the scene from the perspective of the camera. Novel View Synthesis applied at the level of outdoor scenes is an exciting research direction that can enable users to explore outdoor areas with greater fidelity as they are no longer confined to the camera's captured views.

A recent development in generating novel views of scenes are neural radiance fields (NeRF)~\cite{mildenhall2021nerf} that allows a complex 3D scene to be compactly represented using a coordinate-based network implemented with multilayer perceptrons (MLP). Using volumetric rendering, NeRF is able to generate photo-realistic novel views and it has inspired many subsequent works~\cite{lin2021barf, nerfw, inerf, meng2021gnerf, dsnerf, urbanradiancefields, lunerf, nopenerf} that try to address some of its shortcomings. However, despite the drawbacks that have been tackled by the above works, these methods still require sampling the volume. Thus, they share the limitations of the volumetric approach such as being limited to small scenes with strong parallax support~\cite{ost2022neural}. As a result, applying them to outdoor scenes is an inherently difficult problem. The work of Ost et al.~\cite{ost2022neural} proposes to extend the ability to represent large scenes with neural point light fields (NPLF). However, in realistic scenarios, outdoor scenes tend to have multiple inconsistencies between captured RGB frames, particularly due to transient or moving objects. Additionally street scenes normally consist of longer videos and are thus sensitive to camera pose drifts. To generate high-quality photo-realistic synthetic views, camera poses need to be refined as well. These two major issues have been partly addressed by methods like~\cite{ost2022neural, blocknerf, urbanradiancefields, li2019aads, yang2020surfelgan}. However, none of these methods address them jointly. 
In our study, we address these significant challenges concurrently. Our method generates high-fidelity street view imagery while autonomously managing dynamic moving objects, eliminating the need for manual annotations. Additionally, we simultaneously refine the initial camera poses to enhance the quality of the renderings. To summarize, our key contributions are:
\begin{itemize}
    \item We incorporate novel view synthesis with dynamic object erasing which removes artifacts created by inconsistent frames in urban scenes. 
    \item We propose a voting scheme for dynamic object detection to achieve consistent classification of moving objects.
    \item During training, we jointly refine camera poses and demonstrate the robustness of our method to substantial camera pose noise. As a result, image quality is elevated with the increased accuracy of camera poses.
    \item We validate our method on real-world urban scenes in a variety of conditions, with varying amounts of pedestrian and vehicle traffic, and the experiments show that we achieve state-of-the-art results.
\end{itemize}

\section{Related Work}
\paragraph{Outdoor Novel View Synthesis} The explosion in popularity of coordinate-based networks can be attributed to their efficiency and photo-realism when synthesizing novel views of small, object-centric scenes, but they do not naively scale well to larger scenes~\cite{blocknerf}. Block-NeRF~\cite{blocknerf} proposes to break down a larger radiance field network into smaller compact radiance fields, which are then composited together to render a given view. However, training multiple NeRF models is computationally demanding and difficult with consumer hardware. A more efficient method is to embed local radiance functions or light fields on proxy geometry~\cite{ost2022neural, yang2020surfelgan}. Embedding light fields on point clouds~\cite{ost2022neural} is both more efficient and better for outdoor scenes as it does not require strong parallax as in volumetric methods and requires a single evaluation per ray. These methods however neglect the presence of dynamic objects in outdoor scenes that introduce inconsistencies between image frames leading to artifacts in rendered images, see~\cref{fig:teaser}. Some methods address this issue by modeling the scene as the composition of static and transient objects~\cite{nerfw}, by masking every potential moving object~\cite{blocknerf, li2019aads}, or by masking out human annotated dynamic objects~\cite{yang2020surfelgan}. We show that by using motion detection and masking out dynamic objects during training, our method achieves much better results in outdoor scenes without the need to lose any information due to masking out every static 
object, such as parked vehicles, or needing to have any ground truth annotations for dynamic objects.   
\vspace{-0.4cm}
\paragraph{View Synthesis with Pose Refinement} While novel view synthesis methods are self-supervised, most of them assume the presence of posed images~\cite{meng2021gnerf} \ie the camera transformations between different images are known. Accurate camera poses are essential to correctly reconstruct a scene and to ensure view consistency and these are generally estimated using structure from motion (SfM)~\cite{nerf--} such as COLMAP~\cite{colmap} or are obtained using motion-capture systems when the ground truth camera poses are not available. However, SfM works by tracking a set of keypoints across multiple views to estimate the relative poses. If these keypoints are located on moving objects such as cars and pedestrians, the relative motion between the camera and the object will lead to view inconsistencies~\cite{li2021unsupervised} and thus inaccurate camera poses. Most of the previous view synthesis methods work well when the poses are accurate but inaccuracies in the camera pose can often degrade photorealism~\cite{lunerf}. Thus, joint optimization of the scene representation and the camera poses using the gradients obtained from an image loss~\cite{lin2021barf, meng2021gnerf, nerf--} is commonly used to alleviate this issue. To prevent the pose refinement from being stuck in sub-optimal minima, BARF~\cite{lin2021barf} proposed to use a coarse-to-fine optimization process by using a dynamic low-pass filter on the positional encodings. Assuming a roughly estimated radiance field is available, some methods~\cite{inerf, meng2021gnerf} train an inversion network to estimate the camera poses which are then further refined. Relative poses between adjacent frames are also used to inform the global pose optimization -~\cite{nopenerf} uses geometry cues from monocular depth estimation to estimate relative poses between nearby frames and~\cite{lunerf} computes the local relative poses in mini scenes of a few frames before performing a global optimization for the whole scene. While these methods show improvements to NeRF, they have not been exploited in outdoor scenes with moving objects.
\vspace{-0.4cm}
\paragraph{Neural Point Light Fields} Neural Point Light Fields (NPLF) encodes scene features on top of a captured point cloud to represent a light field of the scene. Similar to coordinate-based methods, neural point light field (NPLF) generates synthetic images by shooting rays per pixel and rendering the color according to the features of the ray. However, NPLF makes use of the corresponding point cloud of the scene such that the ray features are not extracted by sampling the volume hundreds of times per ray, but is rather obtained by aggregating the features of the points close to each ray. As all the required features of the scene are encoded in the light field per point in the point cloud, there is no requirement for volume sampling. This makes NPLF much more efficient at representing large scenes as compared to purely implicit representations like neural radiance fields.
\vspace{-0.4cm}
\paragraph{Dynamic Objects Detection}
Handling moving objects is particularly important in outdoor scene reconstruction and view synthesis because they tend to create ghostly artifacts during rendering if the method fails to detect them. Detecting moving objects is a challenging problem; the goal is to separate the transient foreground from the static background~\cite{giraldo2021graph}. Some of the earliest research to detect moving objects was change detection, where the background is subtracted from scenes to get a foreground mask~\cite{st2014subsense, sedky2014spectral}. While these methods work well when the camera is static, in the case of view synthesis for street views, the camera is typically moving through the scene, and therefore the assumption of a static background no longer holds. Other approaches include modeling the scene using graphs~\cite{giraldo2021graph, giraldo2021graphbgs} or by using additional information such as disparity maps~\cite{talukder2004real} or optical flow~\cite{lo2022optical, huang2018optical}. We exploit a combination of optical flow, which encodes the relative motion between the observer and the scene~\cite{dosovitskiy2015flownet, teed2020raft, kong2021fastflownet}, binary classification, and object tracking to detect and remove moving objects from the training process.       

In this work, we show that by properly detecting moving objects and excluding the corresponding pixels during the ray-marching-based training phase, we can obtain novel views of outdoor scenes that do not have artifacts caused by transient objects. Moreover, in practical scenarios, accurate camera poses are often absent in outdoor scenes. We demonstrate the impact of noisy camera poses on the photo-realism of these scenes. We propose to jointly optimize camera poses along with the view synthesis model. By enhancing the accuracy of camera poses, we can achieve better rendering results.


\section{Method}
\begin{figure*}[t!]
    \centering
    \includegraphics[height=4cm]{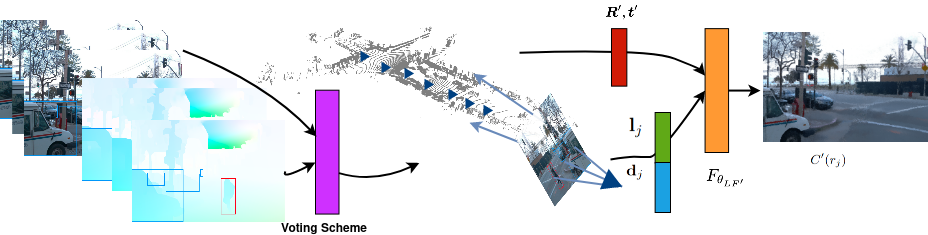}
    \caption{Pipeline of our method: We extract the optical flow of the scene and the object tracking information for each frame in the scene. These are then used by our motion detection network to classify objects as moving or static; our voting scheme ensures that these predictions remain consistent throughout the scene. We then shoot rays through the areas of the images that do not contain any moving objects and extract features for the ray. An MLP then predicts the color of the ray. We update the features, the weights of the MLP, and the camera poses jointly in a self-supervised way.} 
    \label{fig:pipeline}
    \vspace{-0.3cm}
\end{figure*}
\subsection{Neural Point Light Fields}
Inspired by~\cite{ost2022neural}, we use NPLF to represent the scene features. Given a sequence of RGB images $\{\I_1, \I_2, \dots, \I_\mathcal{V}\}_n$ of a scene and their corresponding depth image or LiDAR image $\{\vt{D}_1, \vt{D}_2, \dots, \vt{D}_\mathcal{V}\}$. We first merge these images into a point cloud $\mathcal{P} = \{\vt{x}_1, \vt{x}_2, \dots, \vt{x}_\mathcal{N}\}_n$. We extract features for each point $\vt{x}_k \in \mathcal{P}$ and concatenate them with their locations to obtain per-point feature vectors $\vt{l}_k \in \mathbb{R}^{6 \times f_n}$, where $f_n$ is the feature length. $\vt{l}_k$ has $6$ in the first dimension because an input point cloud is converted to sparse depth images by projecting onto six planes of a canonical cube, and these depth maps are then processed by a convolution neural network to extract per-pixel features. For a ray $\vt{r}_j$ shot through the point cloud, we aggregate the point features of the $k$ nearest points with an attention mechanism to obtain a light-field descriptor $\vt{l}_j$ for this ray. An MLP $F_{\theta_{LF}}$ then predicts the color for the ray $\vt{r}_j$ as $C(\vt{r}_j) = F_{\theta_{LF}}(\vt{d}_j \oplus \vt{l}_j)$, where $\vt{d}_j$ is the viewing direction. The method is trained by minimizing the error between the original color of the RGB image and the predicted color. For the sake of brevity, we refer to the original paper~\cite{ost2022neural} for more details.

\subsection{Moving Object Detection}\label{sec:voting_scheme}
We use a moving object detection algorithm based on~\cite{lo2022optical} with an additional voting scheme to detect moving objects in the RGB images. Using an off-the-shelf object tracking system, we extract object IDs and bounding boxes for each detected vehicle and pedestrian in the image sequence $\{\I_n\}$. Next, we extract the optical flow for the images and crop them to the bounding boxes of the detected objects. We rescale the cropped optical flows for each object while maintaining their aspect ratios. Then, we use a convolutional classification network to predict if the object localized at the given bounding box is moving. Additionally, we employ a voting scheme to reduce inconsistencies in motion prediction that may be caused by incorrect optical field computation or the inconsistencies introduced by ego-motion. In frame $j$ where the object with instance $i$ appears, we compute the motion score $m_j^i\in \{0,1\}$, where $1$ and $0$ denote moving and non-moving objects respectively. Thus, each object has a sequence of motion labels $\{m^i_n\}_n$ (out side $n$ means iterate over $n$) indicating their motion statuses over frames. Finally, the motion status $M^i$ of an object instance $i$ across the scene is set as
\begin{equation}
  M^{i} = \begin{cases}
        1 \text{ if } \med(\{m^i_n\}_n) \geq 0.5 \,, \\
        0 \text{ otherwise }\,,
          \end{cases}
  \label{eq:voting_scheme}
\end{equation}
where $\med(\{m^i_n\}_n)$ is the median of the motion labels for object $i$ in the sequence $\{m^i_n\}_n$. If an instance object is labeled as $1$, we denote this object as moving over the entire sequence.

\subsection{Masked NPLF} \label{sec:masking}
\begin{wrapfigure}{r}{3cm}
    \centering
    \includegraphics[height=\linewidth, angle=270]{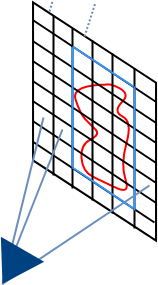}
    \caption{}
    \label{fig:mask_ray}
\end{wrapfigure}
We use the motion labels $M^i$, to mask out the transient areas for each frame $\I_n$. For each pixel in frame $n$, we select the pixels that do not contain any moving objects to perform ray marching. For example, in~\cref{fig:mask_ray}, the red object is classified as a moving object and the blue color box indicates the bounding box of this object. When shooting a ray through the pixels of the image, we do not shoot rays that go through the pixels inside the blue bounding box. Denote
$\mathcal{R^{\prime}}$ as the set of rays that are cast from the camera center to the non-masked pixels only. This allows us to retain the information from static vehicles unlike previous masking-based approaches, which mask out all instances of commonly transient objects. Additionally, we reduce the uncertainty introduced by objects that are in motion, which is a very common feature of outdoor scenes. At inference time, we do not consider the mask and instead shoot rays through the entire pixel grid.

\subsection{Pose Refinement}
To solve the aforementioned inaccurate camera pose problem, we jointly refine the camera poses with the point light field to account for these potential inaccuracies.
We use the logarithmic representation of the rotation matrix such that the direction and the $l2$ norm of the rotation vector $\mat{R} \in \mathbb{R}^{3}$ represents the axis and magnitude of rotation of the camera in the world-to-camera frame respectively. The translation vector $\vt{t} \in \mathbb{R}^{3}$ represents the location of the camera in the world-to-camera frame. We initialize $\delta \mat{R}$ and $\delta \vt{t}$ as zero-vectors and condition the neural point light field on the refined rotation $\mat{R}^{\prime} = \mat{R} + \delta \mat{R}$ and translation $\vt{t}^{\prime} = \vt{t} + \delta \vt{t}$. We use a weighted positional encoding for the camera pose variables to help the pose converge. We denote this encoding as $\phi: \mathbb{R}^3\to\mathbb{R}^{(2L+1)\times3}$ where $L$ is order of the frequency bases such that 
\begin{align}
    \phi(\vt{x}) = [\dots, \omega_i(L)\gamma_i(\vt{x}), \omega_{i+1}(L)\gamma_{i+1}(\vt{x})\dots] \,,
\end{align}
where the $i$-th frequency encoding $\gamma_{i}$ is 
\begin{equation}
    \gamma_i(\vt{x}) = [\cos(2^i\pi\vt{x}), \sin(2^i\pi\vt{x})]
\end{equation}
We adopt a frequency filter, similar to~\cite{lin2021barf} to gradually introduce the high frequency information of the input resulting in a coarse-to-fine optimization scheme. The smoother starting frequencies make it easier to align the camera poses in the beginning~\cite{lin2021barf}. Once the poses are somewhat correct, the higher frequencies can then be used to learn a high-fidelity scene representation. The weight $\omega_i$ is defined as
\begin{equation}
    \omega_i(L) = \begin{cases} 0\,,  ~~ \text{if} ~~ \alpha < i \\
    \frac{1 - \cos((\alpha - i)\pi)}{2}\,, ~~ \text{if} ~~ 0 \leq \alpha - i < 1 \\
    1\,,  ~~ \text{if} ~~  \alpha - i \geq 1 \end{cases}
\end{equation}
Thus, the color $C^{\prime}(\vt{r}_j)$ of a ray $\vt{r}_j$ is given by 
\begin{equation}
    \label{eq:refined_nplf}
    C^{\prime}(\vt{r}_j) = F_{\theta_{LF^{\prime}}}(\phi(\vt{d}_j) \oplus \phi(\vt{l}_j), \mat{R}^{\prime}, \mat{t}^{\prime})
\end{equation}
where $\vt{d}_j$ and $\vt{l}_j$ are the ray direction and the feature vector corresponding to $\vt{r}_j$, $F_{\theta_{LF^{\prime}}}$ is an MLP. \par
The loss function is
\begin{equation}
    \label{eq:refined_loss}
    \mathcal{L}_{m,r} = \sum_{j \in \mathdutchcal{R^{\prime}}} || C^{\prime}(r_j) - C(r_j) ||^{2} 
\end{equation}
and the updates to the camera rotation and translation are optimized simultaneously with the neural point light field.

\section{Experiments}
\paragraph{Experimental Setup} We evaluate our method on the Waymo open dataset~\cite{waymo}. We chose 6 scenes from Waymo which we believe are representative of street view scenes with different numbers of static and moving vehicles and pedestrians. We use the RGB images and the corresponding LiDAR point clouds for each scene. We drop out every 10th frame from the dataset for evaluation and train our method on the remaining frames. The RGB images are rescaled by a factor of 0.125 of their original resolutions for training. 
We perturb the camera poses for each scene with multiplicative noise $\delta \sim \mathcal{N}(0, 0.1)$. We show our results in two parts. The first part is our voting scheme for motion detection and the second part is the novel view reconstruction and extrapolation results for street scenes.  
\begin{figure*}[ht!]
\setlength{\tabcolsep}{2pt}
\newcolumntype{Y}{>{\centering\arraybackslash}p{0.152\textwidth}}
\begin{tabular}{lYYYYYY}
 & Frame $i$ &  Frame $i+1$ &  Frame $i+2$ &  Frame $i$ &  Frame $i+1$ & Frame $i+2$ \\
    \multirow{2}{*}{\rotatebox{90}{w/o voting}} & \includegraphics[width=\linewidth]{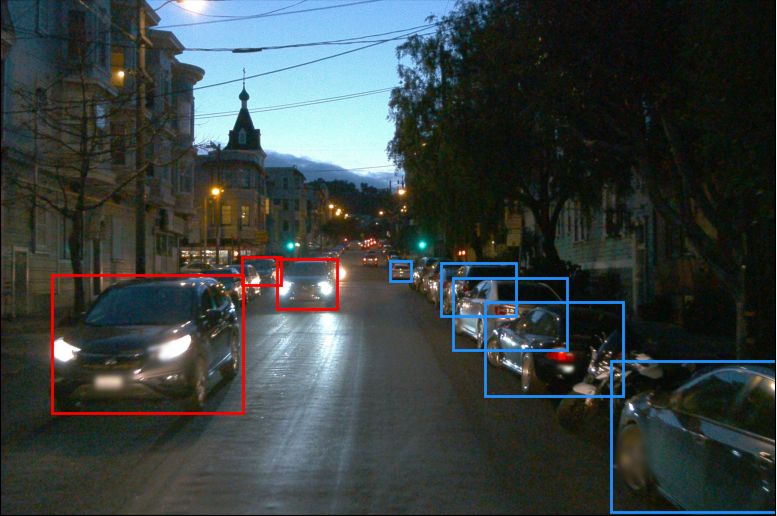} &
    \includegraphics[width=\linewidth]{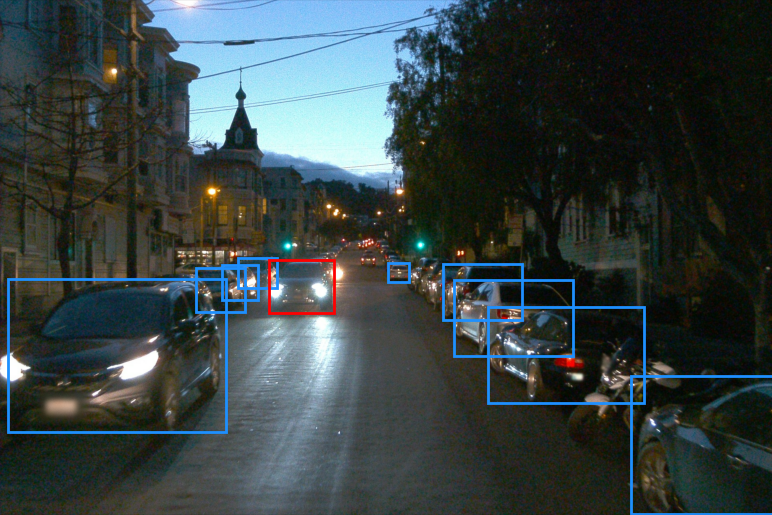} &
    \includegraphics[width=\linewidth]{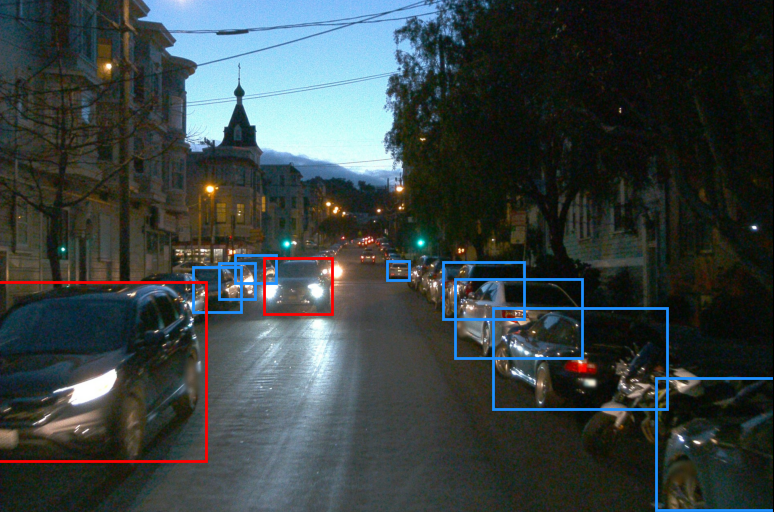} &
    \includegraphics[width=\linewidth]{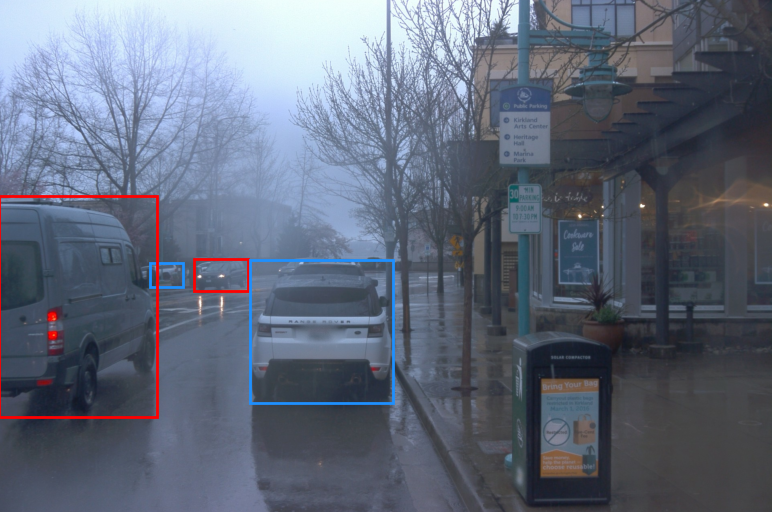} &
    \includegraphics[width=\linewidth]{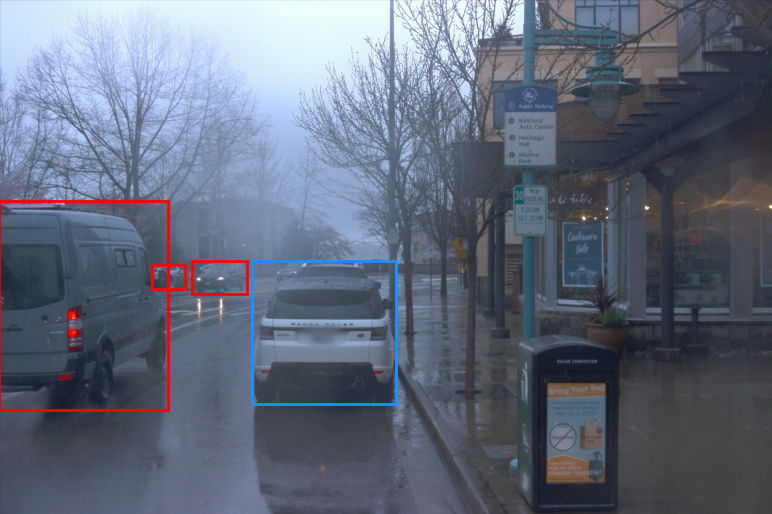} &
    \includegraphics[width=\linewidth]{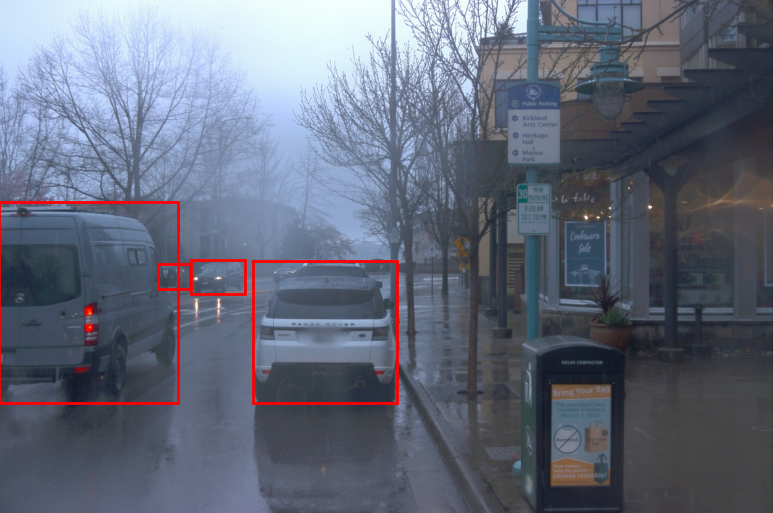} \\
    &
    \includegraphics[width=\linewidth]{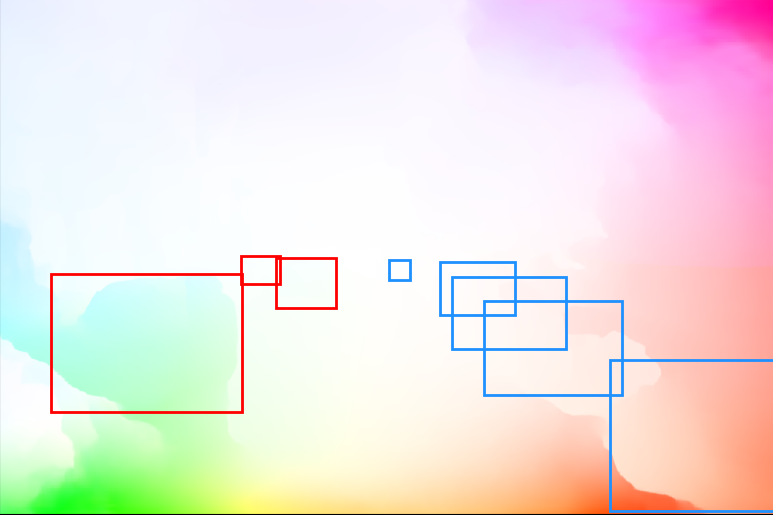} &
    \includegraphics[width=\linewidth]{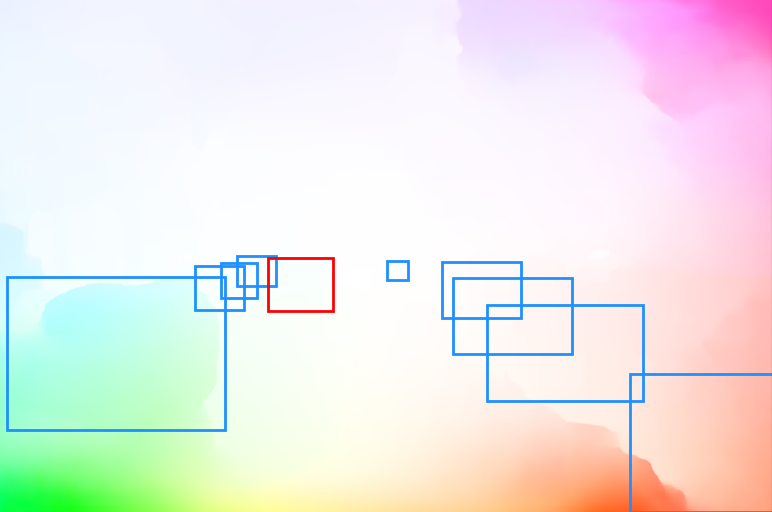} &
    \includegraphics[width=\linewidth]{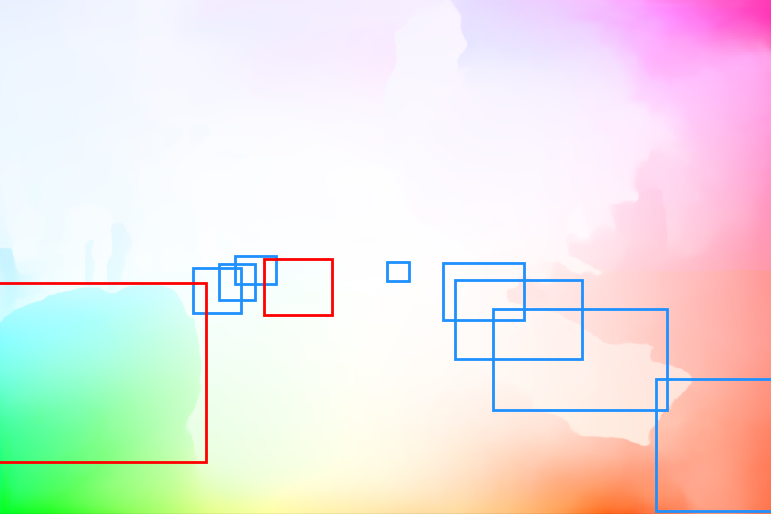} &
    \includegraphics[width=\linewidth]{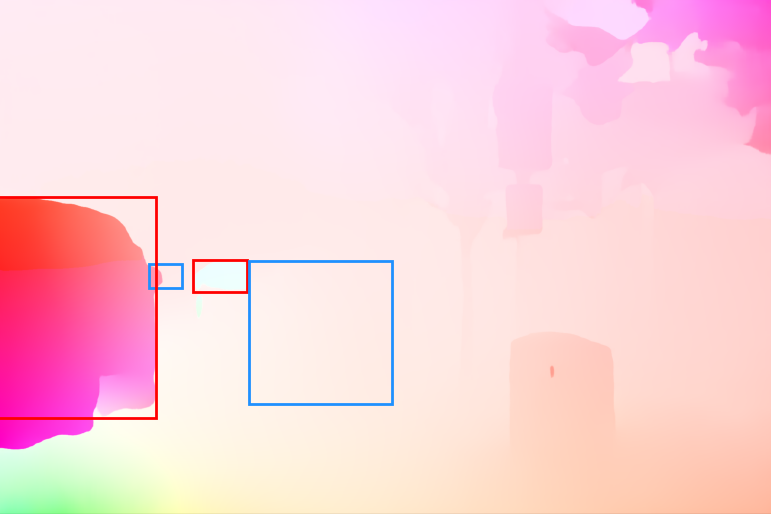} &
    \includegraphics[width=\linewidth]{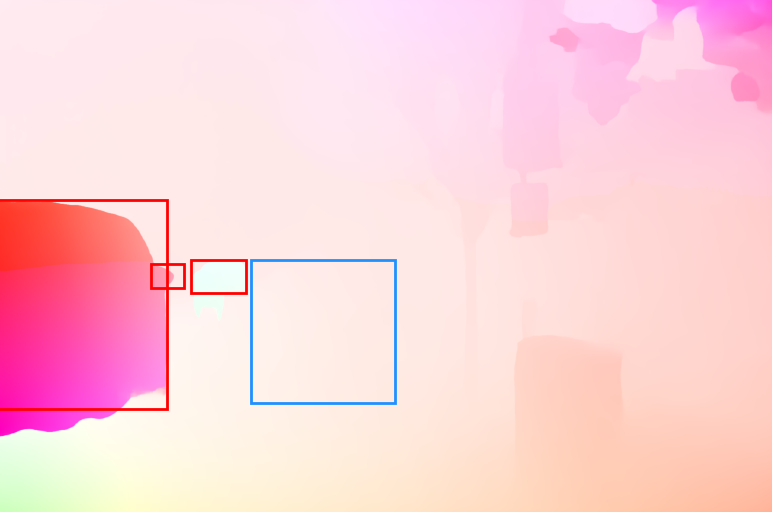} &
    \includegraphics[width=\linewidth]{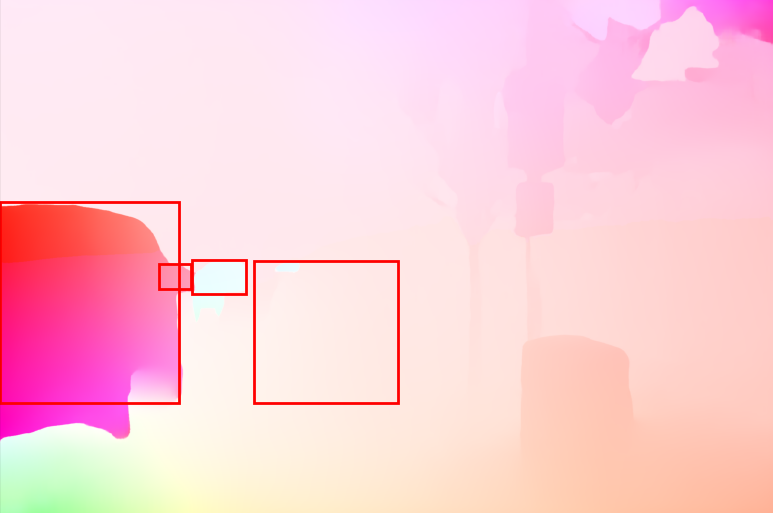} \\
    \multirow{2}{*}{\rotatebox{90}{with voting}}& \includegraphics[width=\linewidth]{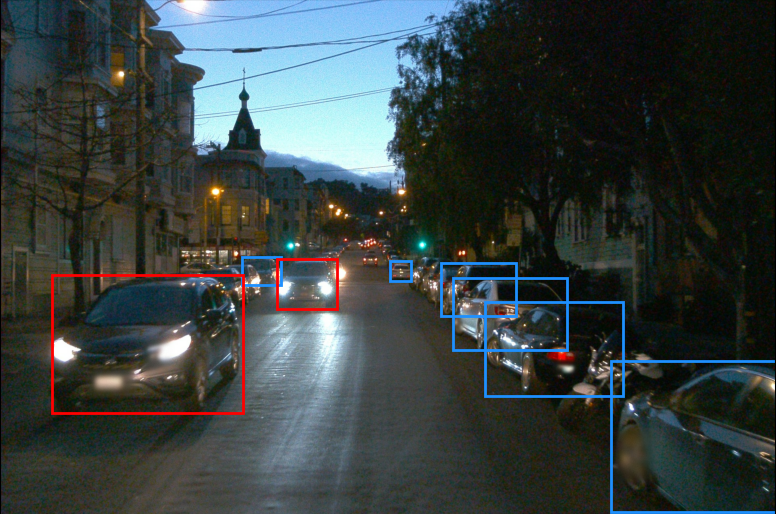} & 
    \includegraphics[width=\linewidth]{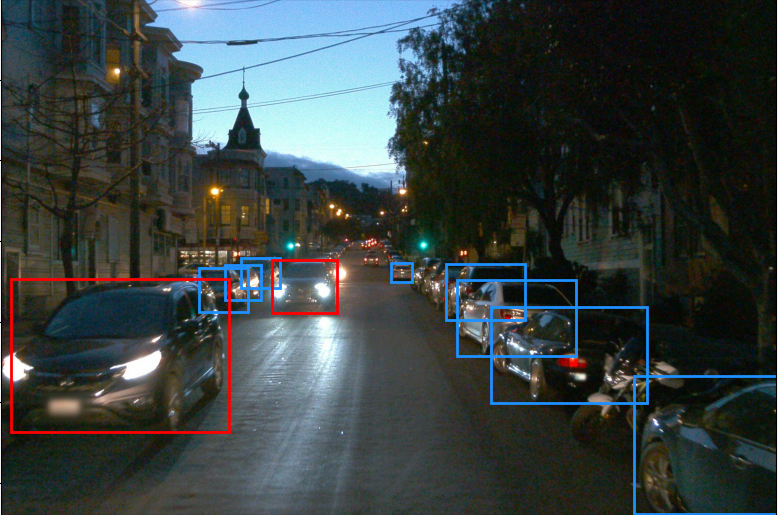} & 
    \includegraphics[width=\linewidth]{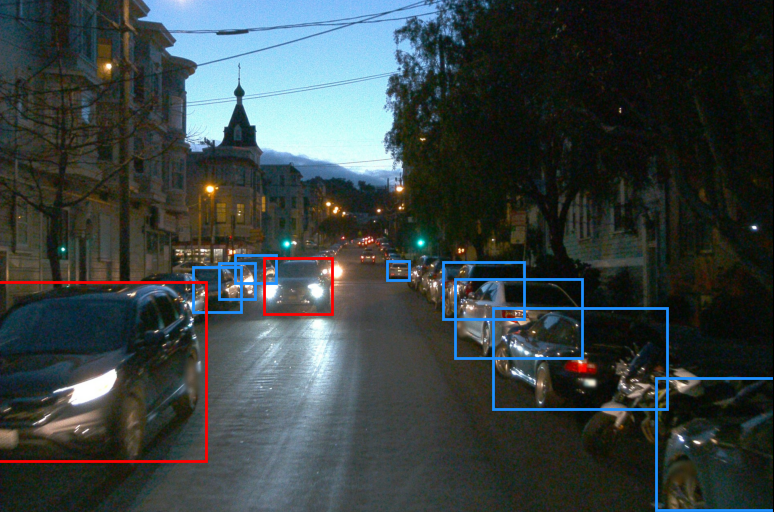} & 
    \includegraphics[width=\linewidth]{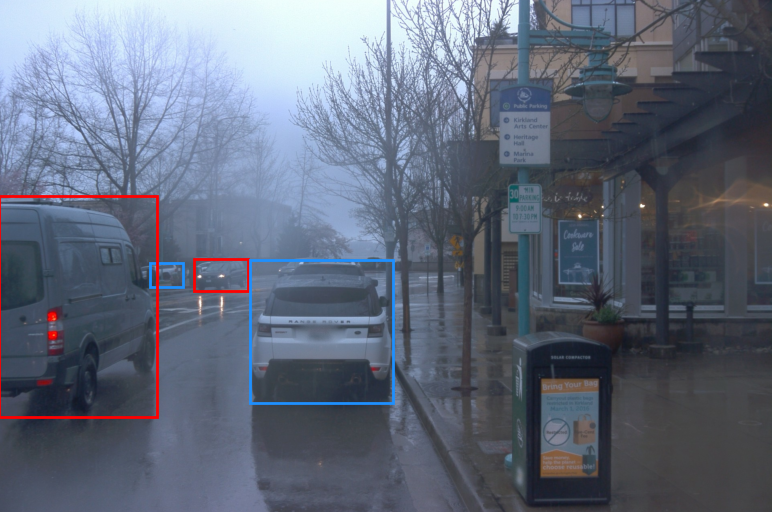} & 
    \includegraphics[width=\linewidth]{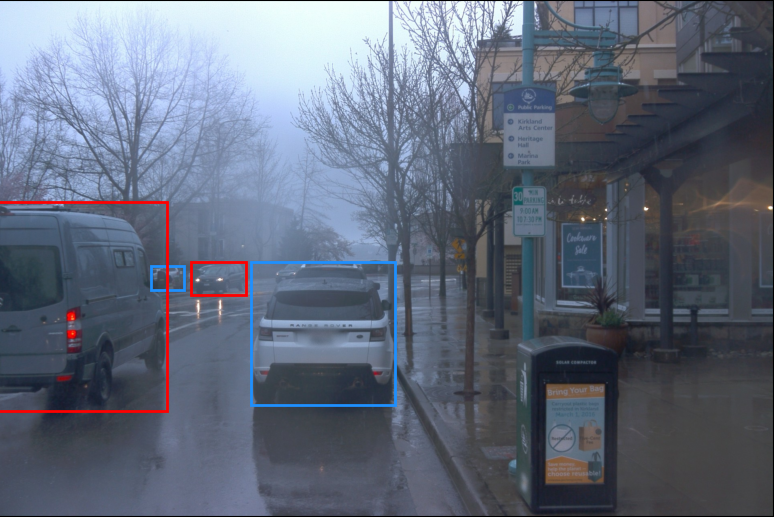} & 
    \includegraphics[width=\linewidth]{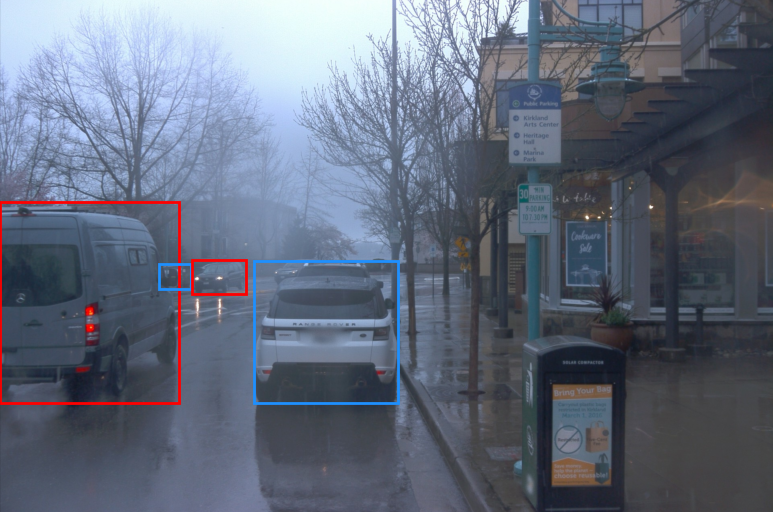} \\
    & \includegraphics[width=\linewidth]{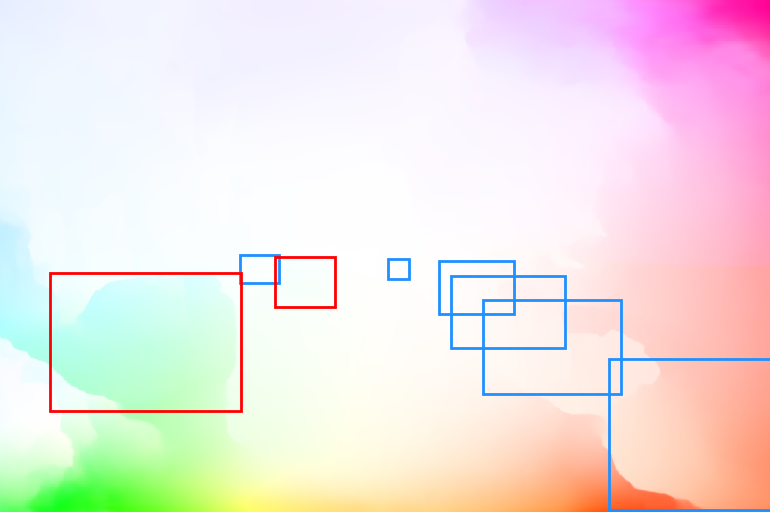} & 
    \includegraphics[width=\linewidth]{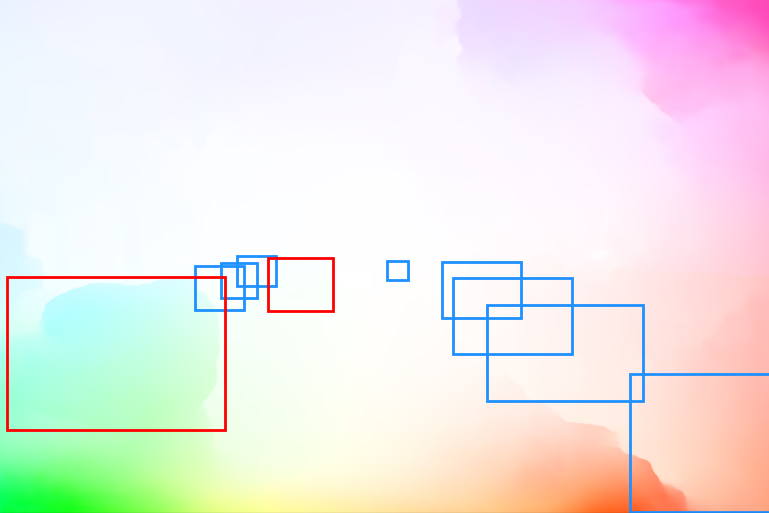} & 
    \includegraphics[width=\linewidth]{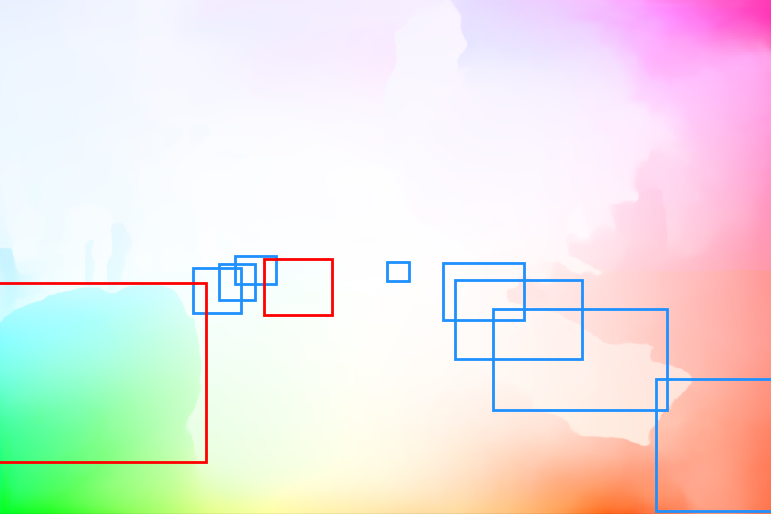} & 
    \includegraphics[width=\linewidth]{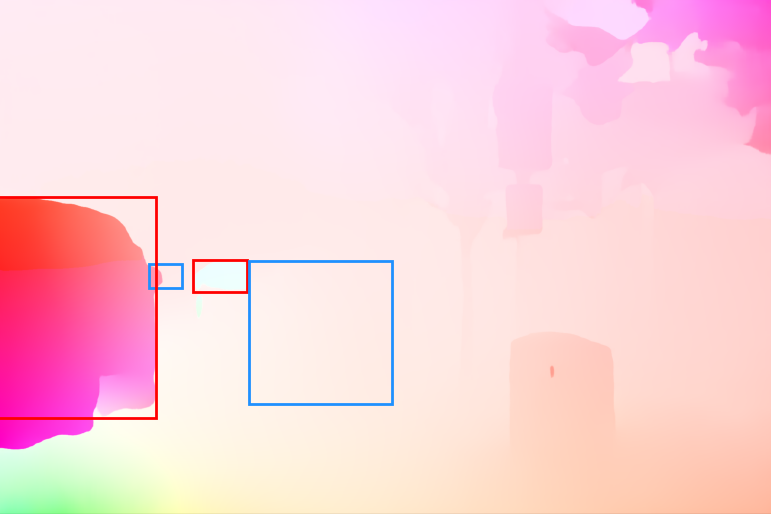} & 
    \includegraphics[width=\linewidth]{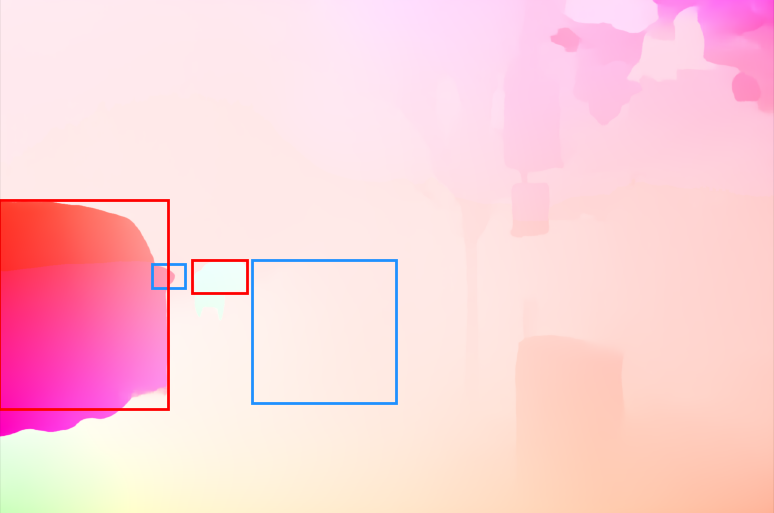} & 
    \includegraphics[width=\linewidth]{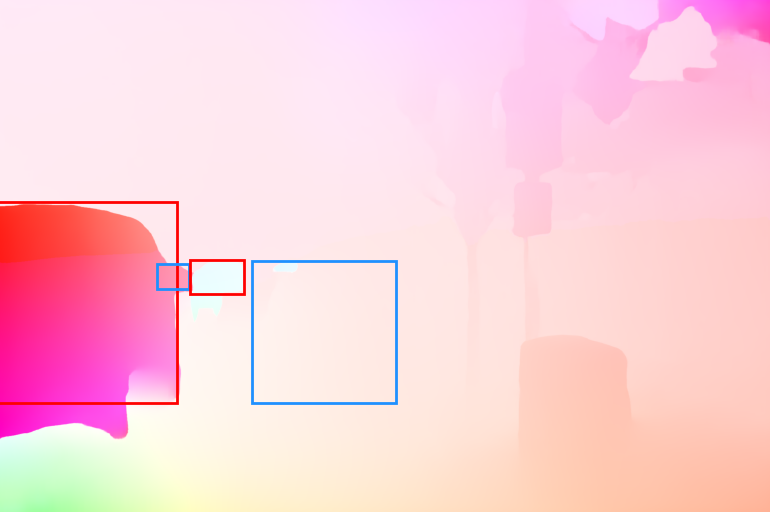}
\end{tabular}
\caption{Motion Detection Results: Our voting scheme ensures temporally consistent dynamic object prediction across a sequence of three consecutive frames. Red bounding boxes indicate moving objects while blue bounding boxes indicate static objects. The first and third rows are the RGB images, the second and last rows are the optical flow of the corresponding images.}    \label{fig:motion_detection_results}
\end{figure*}

\subsection{Motion Detection with Voting Scheme} 
We mask out the moving objects using the method introduced in~\cref{sec:masking} before training. For this, we use RAFT~\cite{teed2020raft} to extract the optical flow maps and YOLO-V8s ~\cite{yolov8_ultralytics} implementation of the ByteTrack algorithm ~\cite{zhang2022bytetrack} to track the objects. We then use the pretrained model from~\cite{lo2022optical} for the initial motion detection outputs. 
\cref{fig:motion_detection_results} shows our motion detection algorithm with and without the voting scheme. We display the motion detection results for three consecutive frames. The blue and red bounding boxes indicate static and moving objects respectively. We can see that in the absence of our voting scheme introduced in~\cref{sec:voting_scheme}, the predicted moving objects can be inconsistent between frames. In the left scene, the moving car in the foreground is classified as parked in the frame $i+i$, and in the right scene, the central parked car is classified as moving in the frame $i+2$. In the second row, we correctly classify the objects as dynamic or static and our voting scheme ensures that these predictions are temporally consistent. Moreover, we evaluate our motion detection algorithm with and without the voting scheme on the KITTI-Motion Dataset compiled by~\cite{smsnet} and report the quantitative evaluation on \cref{tab:motion_metrics}. The higher IoU of the segmentation masks for dynamic objects computed with the voting scheme further shows that incorporating the object tracking information in motion detection leads to better and more consistent results. 
\begin{table}
  \centering
  \begin{tabular}{@{}lc@{}}
    \toprule
     Method & IOU $\uparrow$ \\  
    \midrule
    Without Voting Scheme & 0.584 \\
    With Voting Scheme & \textbf{0.691}\\
    \bottomrule
  \end{tabular}
  \caption{Intersection over Union (IoU) of the segmentation masks for dynamic objects computed with and without the voting scheme. Our voting scheme ensures that dynamic object predictions are consistent and leads to higher IoU.}
\vspace*{-0.4cm}
  \label{tab:motion_metrics}
\end{table}

\begin{figure*}[ht!]
\setlength{\tabcolsep}{2pt}
\newcolumntype{Y}{>{\centering\arraybackslash}p{0.19\textwidth}}
\begin{tabular}{YYYYY}
  Ground truth & Nerf-W~\cite{nerfw} &  Nope-NeRF~\cite{nopenerf} &  NPLF~\cite{ost2022neural} &  Ours \\
  \includegraphics[width=\linewidth]{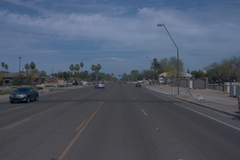} &
  \includegraphics[width=\linewidth]{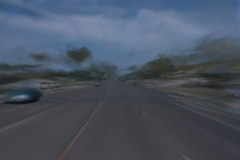} &
    \includegraphics[width=\linewidth]{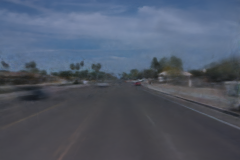} &
    \includegraphics[width=\linewidth]{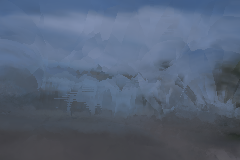} &
    \includegraphics[width=\linewidth]{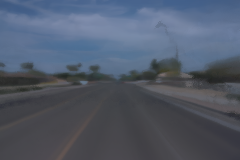} \\
    
    \includegraphics[width=\linewidth]{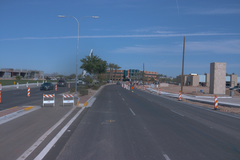} &
  \includegraphics[width=\linewidth]{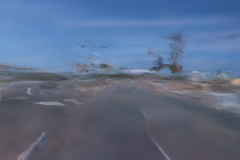} &
    \includegraphics[width=\linewidth]{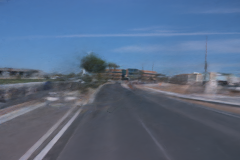} &
    \includegraphics[width=\linewidth]{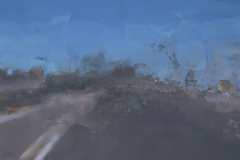} &
    \includegraphics[width=\linewidth]{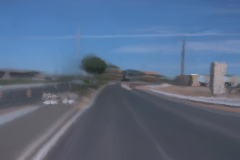} \\
    
    \includegraphics[width=\linewidth]{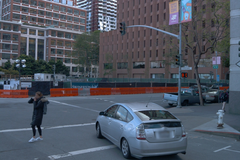} &
    \includegraphics[width=\linewidth]{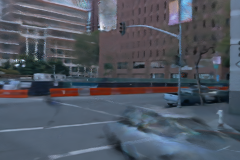} &
    \includegraphics[width=\linewidth]{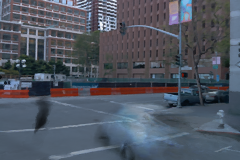} &
    \includegraphics[width=\linewidth]{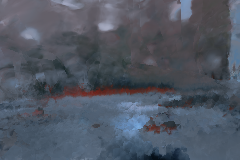} &
    \includegraphics[width=\linewidth]{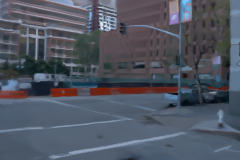} 
 \end{tabular}
\caption{Reconstructions: We compare the reconstruction capability of our method against Neural Point Light Fields~\cite{ost2022neural}, NeRF in the Wild~\cite{nerfw}, and Nope-NeRF~\cite{nopenerf}. All methods were trained on the Waymo Open Dataset \cite{waymo} with artificially added noise to the camera poses to simulate real-world use cases. We query the methods to reconstruct previously seen views. Our method successfully renders large urban scenes without any ghosting artifacts.}    \label{fig:recon_views}
\end{figure*}
\begin{figure*}[t!]
\setlength{\tabcolsep}{2pt}
\newcolumntype{Y}{>{\centering\arraybackslash}p{0.19\textwidth}}
\begin{tabular}{YYYYY}
  Ground truth & Nerf-W~\cite{nerfw} &  Nope-NeRF~\cite{nopenerf} &  NPLF~\cite{ost2022neural} &  Ours \\
  \includegraphics[width=\linewidth]{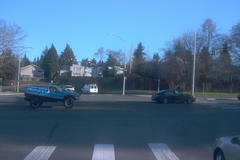} &
  \includegraphics[width=\linewidth]{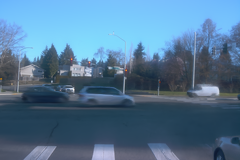} &
    \includegraphics[width=\linewidth]{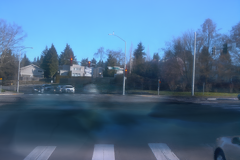} &
    \includegraphics[width=\linewidth]{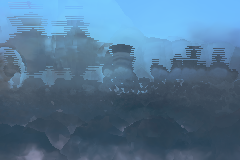} &
    \includegraphics[width=\linewidth]{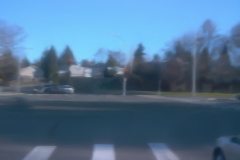} \\
    
    \includegraphics[width=\linewidth]{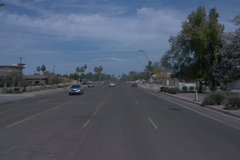} &
  \includegraphics[width=\linewidth]{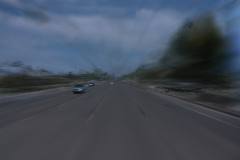} &
    \includegraphics[width=\linewidth]{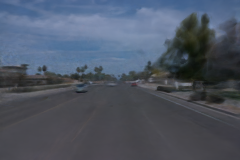} &
    \includegraphics[width=\linewidth]{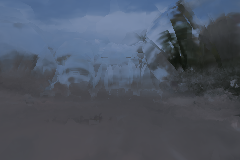} &
    \includegraphics[width=\linewidth]{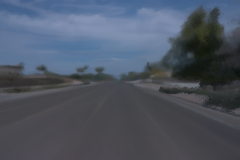} \\
    
    \includegraphics[width=\linewidth]{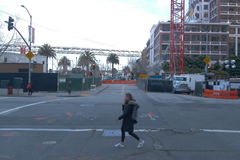} &
  \includegraphics[width=\linewidth]{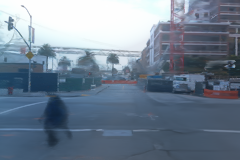} &
    \includegraphics[width=\linewidth]{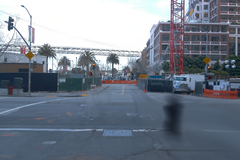} &
    \includegraphics[width=\linewidth]{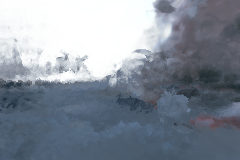} &
    \includegraphics[width=\linewidth]{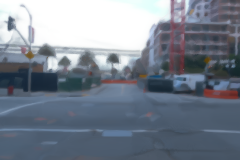} \\
 \end{tabular}
\caption{Novel View Synthesis: We compare the capability of our method in view synthesis against Neural Point Light Fields~\cite{ost2022neural}, NeRF in the Wild~\cite{nerfw}, and Nope-NeRF~\cite{nopenerf}. All methods were trained on the Waymo Open Dataset~\cite{waymo} with artificially added noise to the camera poses to simulate real-world use cases. We render images from interpolated views between the training views, which were held out during training. Our method generates novel views without any ghosting artifacts unlike the compared methods.}    \label{fig:novel_views}
\vspace{-0.35cm}
\end{figure*}
\begin{figure*}[t!]
\setlength{\tabcolsep}{1pt}
\newcolumntype{Y}{>{\centering\arraybackslash}p{0.19\linewidth}}
\begin{tabular}{YYYYY}
  Trajectory  & Nerf-W~\cite{nerfw} &  Nope-NeRF~\cite{nopenerf} & NPLF~\cite{ost2022neural}  &  Ours \\
  \includegraphics[width=\linewidth]{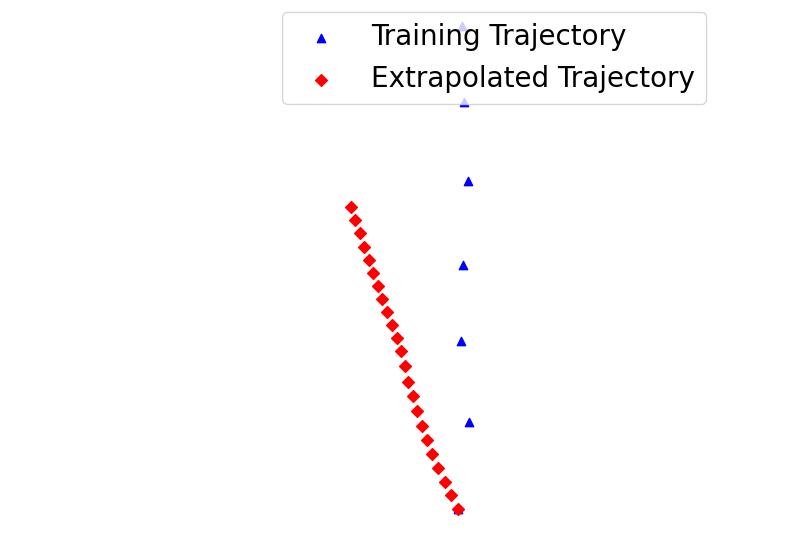} & 
  \includegraphics[width=\linewidth]{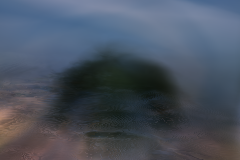} &
    \includegraphics[width=\linewidth]{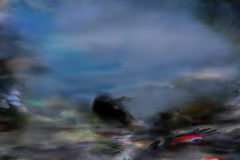} & 
    \includegraphics[width=\linewidth]{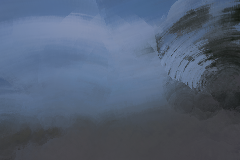} &
    \includegraphics[width=\linewidth]{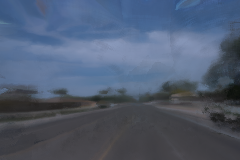} \\
    \includegraphics[width=\linewidth]{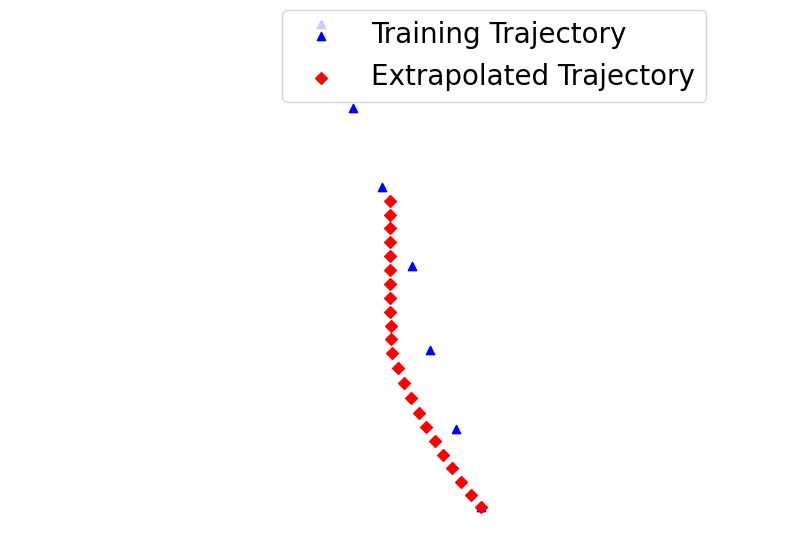} & 
   \includegraphics[width=\linewidth]{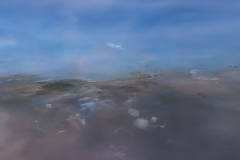} &
    \includegraphics[width=\linewidth]{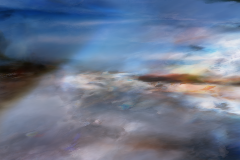} &
    \includegraphics[width=\linewidth]{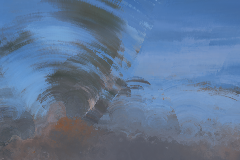}&
    \includegraphics[width=\linewidth]{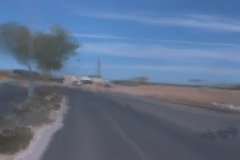} \\
 \end{tabular}
\caption{Trajectory Extrapolation: We present the results of rendering the trained models on camera views that are far away from the training views. The first column shows the extrapolated trajectory vs the training trajectory. Compared to Nerf in the Wild (NeRF-W)~\cite{nerfw}, Nope-NeRF~\cite{nopenerf}, and Neural Point Light Fields~\cite{ost2022neural} trained on the same scenes from Waymo Open Dataset~\cite{waymo}, our method can successfully render clear views whereas the other methods completely collapse.}  
\end{figure*}


\subsection{View Reconstruction}
We evaluate our method's ability to reconstruct the previously seen views (reconstruction) and unseen views (novel view synthesis). We compare qualitatively and quantitatively with 
the baseline method NPLF~\cite{ost2022neural}, as well as previous works NeRF-W~\cite{nerfw} and Nope-Nerf~\cite{nopenerf}. NeRF-W~\cite{nerfw} is developed for outdoor scenes with transient objects, and Nope-NeRF's~\cite{nopenerf} central contribution is pose-free radiance fields. For Nope-NeRF~\cite{nopenerf}, we extract the depth maps using the provided checkpoints for the monocular depth estimator DPT~\cite{dpt}.
\vspace{-0.4cm}
\paragraph{Scene and novel view reconstruction}~\cref{fig:recon_views} shows our scene reconstruction results as well as those of the compared methods. While our method can successfully recover the scene with the dynamic objects, the comparison methods all create ghostly shadows in these areas. Our reconstructed images are clearer than the comparison methods as well.~\cref{fig:novel_views} shows the novel view synthesis results. Here, we render views for the camera poses whose corresponding images are not in the training dataset. NeRF-W attempts to model the transient and static components of the scenes separately but the low parallax support, larger scene sizes, and inaccurate camera poses lead to very blurry renderings. In some cases, where the scene has a small baseline \ie the car moves in a smaller distance, (rows 1 and 3 of~\cref{fig:novel_views}), NeRF-W~\cite{nerfw} overfits a single frame and renders the same result irrespective of the view as can be seen by the presence of cars that are not there in the ground truth view or the blurry pedestrian at the wrong side of the image. NPLF~\cite{ost2022neural} and Nope-NeRF~\cite{nopenerf} do not address the issues of dynamic objects in scenes at all. In all cases, we can see that our method successfully deletes the presence of such objects in the rendered scenes whereas the compared methods have artifacts caused by them. NPLF~\cite{ost2022neural} completely collapses in the absence of accurate poses as it can no longer shoot rays accurately to render a pixel and is therefore unable to represent a light field using the static point cloud. Implicit methods like NeRF-W~\cite{nerfw} can somewhat recover from inaccurate poses when the spatial extent of the scene is not too large (see rows 1 and 3 in~\cref{fig:novel_views}), but the accumulated pose errors over larger scenes lead to poor quality reconstructions (row 1 in~\cref{fig:recon_views} and row 2 in~\cref{fig:novel_views}), and in some cases it completely collapses (row 2 in~\cref{fig:recon_views}) as well. Both Nope-NeRF and our method successfully refine camera poses to ensure that the scene representation is view consistent.\\
For the quantitative evaluation, we compute the SSIM~\cite{ssim}, and LPIPS~\cite{lpips} scores for the rendered scene. As our method replaces moving objects with the background during rendering, we additionally report the mean of the PSNR and the masked PSNR (where only the parts of the image without moving objects are used to compute the score) as \textbf{PSNR\textsubscript{M}}. This helps us to get a good estimate of the models performance without ignoring the presence of artifacts in the areas with moving objects. We note that evaluating our method solely on the basis of the metrics is imperfect, as by removing the dynamic objects all metrics will report a larger variation with the reference image, whereas in the case of the compared methods even the presence of ghostly artifacts will automatically receive better scores than our method. Nevertheless, we receive very similar scores to Nope-NeRF (which performed the best among the compared methods) and even get higher PSNR\textsubscript{M} in novel view synthesis.    
\vspace{-0.4cm}
\paragraph{Trajectory extrapolation} Our method uses point clouds as geometry priors. To prove that the network learns the actual scene geometry structure, instead of only learning the color appearance along the trained camera odometry, we extrapolate the trajectory to drift off from the training dataset. We then render views from this new trajectory which are far away from the training views. This differs from the novel view synthesis results presented in the previous paragraph where the network rendered views that were interpolated on the training trajectory. By doing this  
we are able to evaluate the quality of our scene representation for large scenes and the capability to encode the geometry of the scene. We compare the rendered views for trajectory extrapolation with Nope-NeRF~\cite{nopenerf}, NeRF-W~\cite{nerfw} and NPLF~\cite{ost2022neural}. Neither NeRF-W, NPLF, nor Nope-NeRF are able to extrapolate their scene representations to the new views. On the other hand, our method successfully renders clear images which shows its suitability for large scene representations. 
\begin{table}
  \centering
  \begin{tabular}{@{}lccc@{}}
    \toprule
    \multirow{2}{*}{Method} & \multicolumn{3}{c}{Reconstruction} \\
             \cmidrule{2-4}
             & PSNR\textsubscript{M} $\uparrow$ & SSIM $\uparrow$ & LPIPS $\downarrow$ \\
    \midrule
    Nerf-W ~\cite{nerfw} & 21.326 & 0.826 & 0.353 \\
    Nope-NeRF ~\cite{nopenerf} & \textbf{26.271} & \textbf{0.857} & \textbf{0.241} \\
    NPLF ~\cite{ost2022neural} & 18.658 & 0.750 & 0.546 \\
    Ours & 26.017 & 0.858 & 0.323 \\
    \midrule
    \multirow{2}{*}{Method} & \multicolumn{3}{c}{Novel View Synthesis} \\
             \cmidrule{2-4}
             & PSNR\textsubscript{M} $\uparrow$ & SSIM $\uparrow$ & LPIPS $\downarrow$ \\
    \midrule
    Nerf-W ~\cite{nerfw} & 20.785 & 0.804 & 0.372  \\
    Nope-NeRF ~\cite{nopenerf} & 25.421 & \textbf{0.856} & \textbf{0.223}\\
    NPLF ~\cite{ost2022neural} & 19.151 & 0.752 & 0.540\\
    Ours & \textbf{25.600} & 0.853 & 0.328 \\
    \bottomrule
  \end{tabular}
  \caption{Quantitative Metrics for scene reconstruction and novel view synthesis. We obtain similar scores to Nope-NeRF even though, by removing the moving objects, we anticipate a higher quantitative error between our rendered image and the ground truth reference images.}
  \label{tab:quantitative_metrics}
  \vspace{-0.4cm}
\end{table}

\begin{figure}[t!]
\setlength{\tabcolsep}{1pt}
\newcolumntype{Y}{>{\centering\arraybackslash}p{0.45\linewidth}}
    \centering
    \begin{tabular}{YY}
        \includegraphics[width=\linewidth]{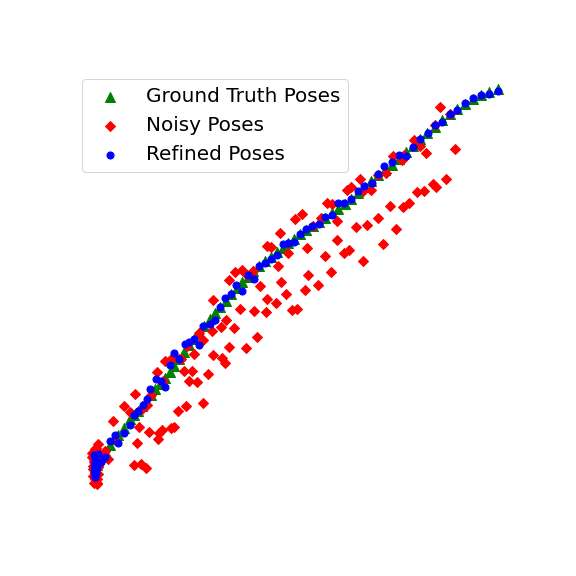} &
        \includegraphics[width=\linewidth]{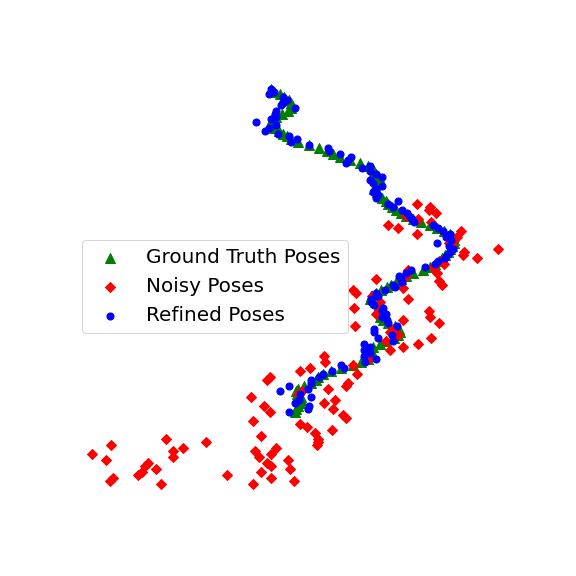} \\
    \end{tabular}
    \caption{We present a visualization of the noisy camera trajectories, the ground truth trajectories, and the refined trajectories. Our pose refinement method successfully recovers accurate poses from very noisy initalizations of the pose.}
    \label{fig:poses}
\end{figure}

\begin{figure}[t!]
\setlength{\tabcolsep}{1pt}
\newcolumntype{Y}{>{\centering\arraybackslash}p{0.22\linewidth}}
    \centering
    \begin{tabular}{cYYYY}
        \rotatebox{90}{recon} & \includegraphics[width=\linewidth]{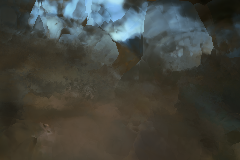} &
        \includegraphics[width=\linewidth]{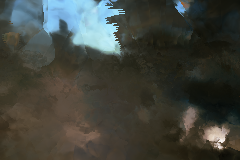} & 
        \includegraphics[width=\linewidth]{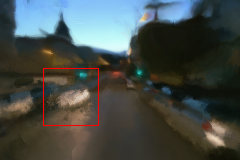} & 
        \includegraphics[width=\linewidth]{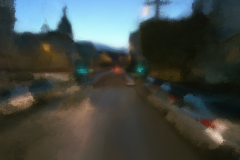} \\
        \rotatebox{90}{novel} &   \includegraphics[width=\linewidth]{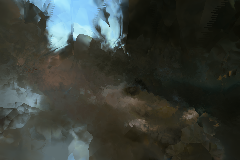} &
        \includegraphics[width=\linewidth]{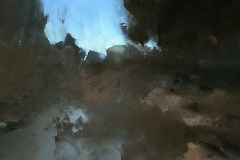} & 
        \includegraphics[width=\linewidth]{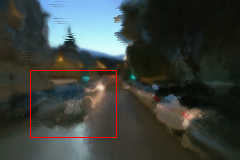} & 
        \includegraphics[width=\linewidth]{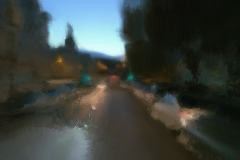}\\
        \rotatebox{90}{recon} &
        \includegraphics[width=\linewidth]{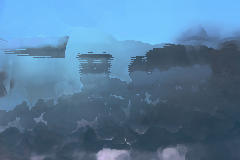} &
        \includegraphics[width=\linewidth]{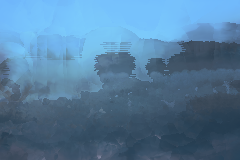} & 
        \includegraphics[width=\linewidth]{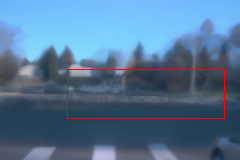} & 
        \includegraphics[width=\linewidth]{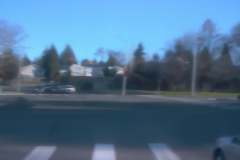}\\
        \rotatebox{90}{novel} &  
        \includegraphics[width=\linewidth]{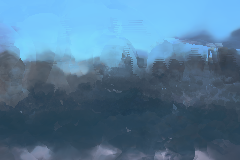} &
        \includegraphics[width=\linewidth]{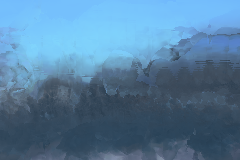} & 
        \includegraphics[width=\linewidth]{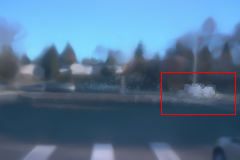} & 
        \includegraphics[width=\linewidth]{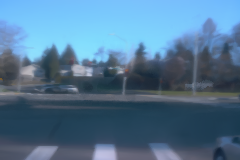}\\
        & Baseline & + Masking & \small{+ Refinement} & Ours \\
    \end{tabular}
    \caption{We compare the baseline NPLF \cite{ost2022neural}, with masking, with refinement, and with both masking and refinement (ours). In the absence of incorrect poses, the results in the first two columns completely collapse, whereas with only refinement and no masking, we get artifacts caused by moving objects (red bounding boxes).}
    \label{fig:ablation}
    \vspace{-0.4cm}
\end{figure}

\vspace{-0.4cm}
\paragraph{Camera pose refinement}
To evaluate the pose refinement step, we compute the absolute trajectory error (ATE) before and after refinement in ~\cref{tab:pose_metrics}. We can see that we successfully recover from very large trajectory errors that accumulate due to inaccuracies of pose through the scene. We also provide visualizations of our pose refinement results in ~\cref{fig:poses}. 
\begin{table}[ht!]
  \centering
  \begin{tabular}{@{}cccc@{}}
    \toprule
     \multirow{2}{*}{Pose} & \multicolumn{3}{c}{ATE (in m) $\downarrow$}\\
     \cmidrule{2-4}
             & Mean & Median & Standard Deviation \\  
    \midrule
    W/o refine & 11.578 & 6.707 & 14.454 \\
    Refined & \textbf{0.010} & \textbf{0.008} & \textbf{0.008} \\
    \bottomrule
  \end{tabular}
  \caption{Average absolute trajectory error (ATE) in metres computed before and after pose refinement. Our joint camera refinement step rectifies inaccuracies in camera pose that can accumulate over long trajectories and affect rendering quality.}
  \label{tab:pose_metrics}
  \vspace{-0.4cm}
\end{table}
\vspace{-0.4cm}
\paragraph{Ablation Study}
To show the effect of dynamic objects masking and camera pose refinement steps, we perform a ablation study on our method in which we evaluate the contribution of each component in our pipeline. We train our pipeline with the following settings: without both (baseline), with only masking (+masking) and with only camera refinement (+refinement) and compare it with the proposed method (+masking + refinement). 
We report the average PSNR\textsubscript{M}, SSIM, and LPIPS scores for both reconstruction and novel view synthesis tasks in~\cref{tab:ablation_metrics} as well as scene rendering results in~\cref{fig:ablation}. The camera refinement step improves the results both quantitatively and qualitatively and is the most crucial part to get successful results. However, only camera refinement without dealing with dynamic objects causes ghostly shadows when generating views as can be seen from the red bounding boxes in ~\cref{fig:ablation}. Quantitatively, we get slightly better PSNR\textsubscript{M} and SSIM scores by adding only refinement because of the reason that we explained in the previous section. 

\begin{table}
  \centering
  \begin{tabular}{@{}lccc@{}}
    \toprule
     Method & PSNR\textsubscript{M} $\uparrow$ & SSIM $\uparrow$ & LPIPS $\downarrow$\\
    \midrule
    NPLF \cite{ost2022neural} & 17.247 & 0.725 & 0.555  \\
    + Masking & 17.414 & 0.730 & 0.554 \\
    + Refinement & \textbf{24.692} & \textbf{0.839} & 0.391 \\
    \midrule
    + Ours & 24.322 & 0.830 & \textbf{0.346}\\
    \bottomrule
  \end{tabular}
  \caption{We compare the quantitative scores averaged over both novel view synthesis and reconstruction for our method against the baseline~\cite{ost2022neural}, baseline with masked out dynamic objects, and baseline with refined poses. Even though removing artifacts penalizes our method quantitatively, our scores are on par or higher than the ablations.}
  \label{tab:ablation_metrics}
  \vspace{-0.4cm}
\end{table}

\section{Conclusion}
\paragraph{Summary} We present an innovative approach for novel view synthesis and camera pose refinement in dynamic outdoor scenes, leveraging neural point light fields and motion detection. 
Our motion detection algorithm identifies and isolates moving objects. The proposed voting scheme efficiently labels dynamic objects consistently over the sequence. This allows our method to generate clear and artifact-free images.
By representing the scene as a light field, our model can generate novel views from a newly generated camera trajectory with high fidelity.  
By jointly refining the camera poses, our method successfully recovers from noisy poses. The accumulation of pose errors would otherwise prevent high-quality renderings. 
By validating on an autonomous driving dataset, we demonstrate that our method achieves state-of-the-art results.
\vspace{-0.4cm}
\paragraph{Limitations and future work} 
Our method uses point clouds as geometry priors which can be easily acquired by depth cameras or LiDAR. However, the quality of the point-cloud can play a role in the rendering quality; a very sparse or inaccurate point cloud may lead to blurry results. Point growing and pruning~\cite{xu2022point} applied to point light fields is an exciting direction for further research. Additionally, we also plan to explore results obtained by masking out the point clouds at the locations of the dynamic objects. 
 

{\small
\bibliographystyle{ieee_fullname}
\bibliography{egbib}

\begin{thebibliography}{10}\itemsep=-1pt

\bibitem{streetview}
Dragomir Anguelov, Carole Dulong, Daniel Filip, Christian Frueh, St{\'e}phane
  Lafon, Richard Lyon, Abhijit Ogale, Luc Vincent, and Josh Weaver.
\newblock Google street view: Capturing the world at street level.
\newblock {\em Computer}, 43(6):32--38, 2010.

\bibitem{nopenerf}
Wenjing Bian, Zirui Wang, Kejie Li, Jia-Wang Bian, and Victor~Adrian
  Prisacariu.
\newblock Nope-nerf: Optimising neural radiance field with no pose prior.
\newblock In {\em Proceedings of the IEEE/CVF Conference on Computer Vision and
  Pattern Recognition}, pages 4160--4169, 2023.

\bibitem{lunerf}
Zezhou Cheng, Carlos Esteves, Varun Jampani, Abhishek Kar, Subhransu Maji, and
  Ameesh Makadia.
\newblock Lu-nerf: Scene and pose estimation by synchronizing local unposed
  nerfs.
\newblock {\em arXiv preprint arXiv:2306.05410}, 2023.

\bibitem{dsnerf}
Kangle Deng, Andrew Liu, Jun-Yan Zhu, and Deva Ramanan.
\newblock Depth-supervised nerf: Fewer views and faster training for free.
\newblock In {\em Proceedings of the IEEE/CVF Conference on Computer Vision and
  Pattern Recognition}, pages 12882--12891, 2022.

\bibitem{dosovitskiy2015flownet}
Alexey Dosovitskiy, Philipp Fischer, Eddy Ilg, Philip Hausser, Caner Hazirbas,
  Vladimir Golkov, Patrick Van Der~Smagt, Daniel Cremers, and Thomas Brox.
\newblock Flownet: Learning optical flow with convolutional networks.
\newblock In {\em Proceedings of the IEEE international conference on computer
  vision}, pages 2758--2766, 2015.

\bibitem{giraldo2021graphbgs}
Jhony~H Giraldo and Thierry Bouwmans.
\newblock Graphbgs: Background subtraction via recovery of graph signals.
\newblock In {\em 2020 25th International Conference on Pattern Recognition
  (ICPR)}, pages 6881--6888. IEEE, 2021.

\bibitem{giraldo2021graph}
Jhony~H Giraldo, Sajid Javed, Naoufel Werghi, and Thierry Bouwmans.
\newblock Graph cnn for moving object detection in complex environments from
  unseen videos.
\newblock In {\em Proceedings of the IEEE/CVF International Conference on
  Computer Vision}, pages 225--233, 2021.

\bibitem{huang2018optical}
Junjie Huang, Wei Zou, Jiagang Zhu, and Zheng Zhu.
\newblock Optical flow based real-time moving object detection in unconstrained
  scenes.
\newblock {\em arXiv preprint arXiv:1807.04890}, 2018.

\bibitem{yolov8_ultralytics}
Glenn Jocher, Ayush Chaurasia, and Jing Qiu.
\newblock Ultralytics yolov8, 2023.

\bibitem{adamoptim}
D Kinga, Jimmy~Ba Adam, et~al.
\newblock A method for stochastic optimization.
\newblock In {\em International conference on learning representations (ICLR)},
  volume~5, page~6. San Diego, California;, 2015.

\bibitem{kong2021fastflownet}
Lingtong Kong, Chunhua Shen, and Jie Yang.
\newblock Fastflownet: A lightweight network for fast optical flow estimation.
\newblock In {\em 2021 IEEE International Conference on Robotics and Automation
  (ICRA)}, pages 10310--10316. IEEE, 2021.

\bibitem{li2021unsupervised}
Hanhan Li, Ariel Gordon, Hang Zhao, Vincent Casser, and Anelia Angelova.
\newblock Unsupervised monocular depth learning in dynamic scenes.
\newblock In {\em Conference on Robot Learning}, pages 1908--1917. PMLR, 2021.

\bibitem{li2019aads}
Wei Li, CW Pan, Rong Zhang, JP Ren, YX Ma, Jin Fang, FL Yan, QC Geng, XY Huang,
  HJ Gong, et~al.
\newblock Aads: Augmented autonomous driving simulation using data-driven
  algorithms.
\newblock {\em Science robotics}, 4(28):eaaw0863, 2019.

\bibitem{lin2021barf}
Chen-Hsuan Lin, Wei-Chiu Ma, Antonio Torralba, and Simon Lucey.
\newblock Barf: Bundle-adjusting neural radiance fields.
\newblock In {\em Proceedings of the IEEE/CVF International Conference on
  Computer Vision}, pages 5741--5751, 2021.

\bibitem{lo2022optical}
Ka~Man Lo.
\newblock Optical flow based motion detection for autonomous driving.
\newblock {\em arXiv preprint arXiv:2203.11693}, 2022.

\bibitem{nerfw}
Ricardo Martin-Brualla, Noha Radwan, Mehdi~SM Sajjadi, Jonathan~T Barron,
  Alexey Dosovitskiy, and Daniel Duckworth.
\newblock Nerf in the wild: Neural radiance fields for unconstrained photo
  collections.
\newblock In {\em Proceedings of the IEEE/CVF Conference on Computer Vision and
  Pattern Recognition}, pages 7210--7219, 2021.

\bibitem{meng2021gnerf}
Quan Meng, Anpei Chen, Haimin Luo, Minye Wu, Hao Su, Lan Xu, Xuming He, and
  Jingyi Yu.
\newblock Gnerf: Gan-based neural radiance field without posed camera.
\newblock In {\em Proceedings of the IEEE/CVF International Conference on
  Computer Vision}, pages 6351--6361, 2021.

\bibitem{mildenhall2021nerf}
Ben Mildenhall, Pratul~P Srinivasan, Matthew Tancik, Jonathan~T Barron, Ravi
  Ramamoorthi, and Ren Ng.
\newblock Nerf: Representing scenes as neural radiance fields for view
  synthesis.
\newblock {\em Communications of the ACM}, 65(1):99--106, 2021.

\bibitem{ost2022neural}
Julian Ost, Issam Laradji, Alejandro Newell, Yuval Bahat, and Felix Heide.
\newblock Neural point light fields.
\newblock In {\em Proceedings of the IEEE/CVF Conference on Computer Vision and
  Pattern Recognition}, pages 18419--18429, 2022.

\bibitem{dpt}
Ren{\'e} Ranftl, Alexey Bochkovskiy, and Vladlen Koltun.
\newblock Vision transformers for dense prediction.
\newblock In {\em Proceedings of the IEEE/CVF international conference on
  computer vision}, pages 12179--12188, 2021.

\bibitem{urbanradiancefields}
Konstantinos Rematas, Andrew Liu, Pratul~P Srinivasan, Jonathan~T Barron,
  Andrea Tagliasacchi, Thomas Funkhouser, and Vittorio Ferrari.
\newblock Urban radiance fields.
\newblock In {\em Proceedings of the IEEE/CVF Conference on Computer Vision and
  Pattern Recognition}, pages 12932--12942, 2022.

\bibitem{colmap}
Johannes~Lutz Sch\"{o}nberger and Jan-Michael Frahm.
\newblock Structure-from-motion revisited.
\newblock In {\em Conference on Computer Vision and Pattern Recognition
  (CVPR)}, 2016.

\bibitem{sedky2014spectral}
Mohamed Sedky, Mansour Moniri, and Claude~C Chibelushi.
\newblock Spectral-360: A physics-based technique for change detection.
\newblock In {\em Proceedings of the IEEE Conference on Computer Vision and
  Pattern Recognition Workshops}, pages 399--402, 2014.

\bibitem{st2014subsense}
Pierre-Luc St-Charles, Guillaume-Alexandre Bilodeau, and Robert Bergevin.
\newblock Subsense: A universal change detection method with local adaptive
  sensitivity.
\newblock {\em IEEE Transactions on Image Processing}, 24(1):359--373, 2014.

\bibitem{waymo}
Pei Sun, Henrik Kretzschmar, Xerxes Dotiwalla, Aurelien Chouard, Vijaysai
  Patnaik, Paul Tsui, James Guo, Yin Zhou, Yuning Chai, Benjamin Caine, et~al.
\newblock Scalability in perception for autonomous driving: Waymo open dataset.
\newblock In {\em Proceedings of the IEEE/CVF conference on computer vision and
  pattern recognition}, pages 2446--2454, 2020.

\bibitem{talukder2004real}
Ashit Talukder and Larry Matthies.
\newblock Real-time detection of moving objects from moving vehicles using
  dense stereo and optical flow.
\newblock In {\em 2004 IEEE/RSJ International Conference on Intelligent Robots
  and Systems (IROS)(IEEE Cat. No. 04CH37566)}, volume~4, pages 3718--3725.
  IEEE, 2004.

\bibitem{blocknerf}
Matthew Tancik, Vincent Casser, Xinchen Yan, Sabeek Pradhan, Ben Mildenhall,
  Pratul~P Srinivasan, Jonathan~T Barron, and Henrik Kretzschmar.
\newblock Block-nerf: Scalable large scene neural view synthesis.
\newblock In {\em Proceedings of the IEEE/CVF Conference on Computer Vision and
  Pattern Recognition}, pages 8248--8258, 2022.

\bibitem{teed2020raft}
Zachary Teed and Jia Deng.
\newblock Raft: Recurrent all-pairs field transforms for optical flow.
\newblock In {\em Computer Vision--ECCV 2020: 16th European Conference,
  Glasgow, UK, August 23--28, 2020, Proceedings, Part II 16}, pages 402--419.
  Springer, 2020.

\bibitem{smsnet}
Johan Vertens, Abhinav Valada, and Wolfram Burgard.
\newblock Smsnet: Semantic motion segmentation using deep convolutional neural
  networks.
\newblock In {\em Proc.~of the IEEE Int.~Conf.~on Intelligent Robots and
  Systems (IROS)}, Vancouver, Canada, 2017.

\bibitem{ssim}
Zhou Wang, Eero~P Simoncelli, and Alan~C Bovik.
\newblock Multiscale structural similarity for image quality assessment.
\newblock In {\em The Thrity-Seventh Asilomar Conference on Signals, Systems \&
  Computers, 2003}, volume~2, pages 1398--1402. Ieee, 2003.

\bibitem{nerf--}
Zirui Wang, Shangzhe Wu, Weidi Xie, Min Chen, and Victor~Adrian Prisacariu.
\newblock Nerf--: Neural radiance fields without known camera parameters.
\newblock {\em arXiv preprint arXiv:2102.07064}, 2021.

\bibitem{xu2022point}
Qiangeng Xu, Zexiang Xu, Julien Philip, Sai Bi, Zhixin Shu, Kalyan Sunkavalli,
  and Ulrich Neumann.
\newblock Point-nerf: Point-based neural radiance fields.
\newblock In {\em Proceedings of the IEEE/CVF Conference on Computer Vision and
  Pattern Recognition}, pages 5438--5448, 2022.

\bibitem{yang2020surfelgan}
Zhenpei Yang, Yuning Chai, Dragomir Anguelov, Yin Zhou, Pei Sun, Dumitru Erhan,
  Sean Rafferty, and Henrik Kretzschmar.
\newblock Surfelgan: Synthesizing realistic sensor data for autonomous driving.
\newblock In {\em Proceedings of the IEEE/CVF Conference on Computer Vision and
  Pattern Recognition}, pages 11118--11127, 2020.

\bibitem{inerf}
Lin Yen-Chen, Pete Florence, Jonathan~T Barron, Alberto Rodriguez, Phillip
  Isola, and Tsung-Yi Lin.
\newblock inerf: Inverting neural radiance fields for pose estimation.
\newblock In {\em 2021 IEEE/RSJ International Conference on Intelligent Robots
  and Systems (IROS)}, pages 1323--1330. IEEE, 2021.

\bibitem{lpips}
Richard Zhang, Phillip Isola, Alexei~A Efros, Eli Shechtman, and Oliver Wang.
\newblock The unreasonable effectiveness of deep features as a perceptual
  metric.
\newblock In {\em Proceedings of the IEEE conference on computer vision and
  pattern recognition}, pages 586--595, 2018.

\bibitem{zhang2022bytetrack}
Yifu Zhang, Peize Sun, Yi Jiang, Dongdong Yu, Fucheng Weng, Zehuan Yuan, Ping
  Luo, Wenyu Liu, and Xinggang Wang.
\newblock Bytetrack: Multi-object tracking by associating every detection box.
\newblock In {\em European Conference on Computer Vision}, pages 1--21.
  Springer, 2022.

\end{thebibliography}
}

\onecolumn
\appendix
\renewcommand \thesection{A\arabic{section}}
\renewcommand{\thefigure}{A.\arabic{figure}}
\renewcommand{\thetable}{A.\arabic{table}}


\section{Used Code and Datasets}

\begin{table*}[ht!]
    \centering
    \begin{tabular}{@{}llllp{7cm}l@{}}
    \toprule
         & Name & type & year & link & license \\
    \midrule
    \cite{nopenerf} & Nope-NeRF & code & 2023 & \small{\url{https://github.com/ActiveVisionLab/nope-nerf/}} & MIT licence\\
    \cite{nerfw} & NeRF-W & code & 2021 & \small{\url{https://github.com/kwea123/nerf_pl}} & MIT license \\
    \cite{ost2022neural} & NPLF & code & 2022 & \small{\url{https://github.com/princeton-computational-imaging/neural-point-light-fields}} & MIT license \\
    \cite{teed2020raft} & RAFT & code & 2020 & \small{\url{https://github.com/princeton-vl/RAFT}} & BSD 3-Clause \\
    \cite{lo2022optical} & MotionDetection & code & 2022 & \small{\url{https://github.com/kamanphoebe/MotionDetection}} & - \\
    \cite{waymo} & Waymo & dataset & 2019 & \small{\url{https://waymo.com/open/}} & -\\
\bottomrule
    \end{tabular}
    \vspace{0.25cm}
    \caption{Used datasets and code in our submission, together with reference, link, and license.}
    \label{tab:code_data}
\end{table*}
\begin{table*}[ht!]
  \centering
  \begin{tabular}{@{}llll@{}}
    \toprule
     Dataset & Set & Scene ID & Frames \\
    \midrule    
    Waymo & validation\_0000 & segment-1024360143612057520\_3580\_000\_3600\_000 & 0-80 \\
          & validation\_0000 & segment-10203656353524179475\_7625\_000\_7645\_000 & 0-100 \\
          & validation\_0000 & segment-10448102132863604198\_472\_000\_492\_000 & 0-150 \\
          & validation\_0000 & segment-10837554759555844344\_6525\_000\_6545\_000 & 30-140 \\
          & validation\_0001 & segment-12657584952502228282\_3940\_000\_3960\_000 & 10-100 \\
          & validation\_0001 & segment-14127943473592757944\_2068\_000\_2088\_000 & 0-100 \\
    \bottomrule
  \end{tabular}
    \caption{Selected scenes from Waymo ~\cite{waymo} datasets used in our experiments.}
  \label{tab:scenes}
\end{table*}

\cref{tab:code_data} summarizes the code and datasets we use for evaluation and comparison. Our code and recorded datasets will be made publicly available upon acceptance. \cref{tab:scenes} summarizes the sequence we used for train and evaluate methods.

\section{Training}
Our method is built on top of the original Neural Point Light Fields codebase ~\cite{ost2022neural} and we share most of the architectural decisions. To aid reproducibility and further experimentation, we will publish our source code. \par
To ensure consistency, we train all scenes with the same set of hyper-parameters. To allow our method to be trained on consumer GPUs (4GB VRAM), we train with a batch size of 2 images, and shoot 256 rays for each image. We use the Adam optimizer ~\cite{adamoptim} which is initialized with a learning rate of $10^{-3}$. \par
We find that merging $2m + 1$ pointclouds, where $m = 10$ in the original paper, often leads to very blurry results. 
We therefore set $m = 3$ which we found to have better results empirically. Similarly, we found that the data augmentation strategy proposed by the original paper where the authors randomly choose an adjacent point cloud $P_{i+h, m}$ with $h \in \{-H, H\}$ from the $H$ neighbouring time steps for training, where $H = 10$, also leads to blurry results in our experiments. Therefore, we set $H = 3$ as well.  \par 
For our weighted positional encoding, we set the ending epoch to $2000$ such that $\alpha = \frac{current epoch}{2000} \times L$, where $L$ is the order of the frequency bases. We set $L = 10$ for the ray feature $l_j$ and $L = 4$ for the ray direction $d_j$.

We choose 
6 scenes from Waymo \cite{waymo} that we believe are representative of urban scenes in different lighting and traffic conditions. We provide the scene ids and the chosen frames in \cref{tab:scenes}.

\section{Additional Results}
\section{Tolerance to Noisy Poses}
\begin{figure*}[t!]
\setlength{\tabcolsep}{2pt}
\newcolumntype{Y}{>{\centering\arraybackslash}p{0.19\textwidth}}
\centering
\begin{tabular}{YYYYY}
Reference & $\mathcal{N}(0,0.1)$ & $\mathcal{N}(0,0.3)$ & $\mathcal{N}(0,0.5)$ & $\mathcal{N}(0,1.0)$ \\
\includegraphics[width=\linewidth]{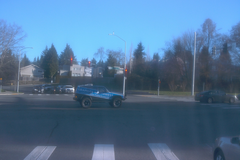} & \includegraphics[width=\linewidth]{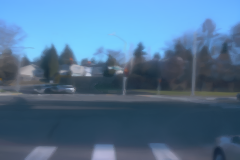} &
\includegraphics[width=\linewidth]{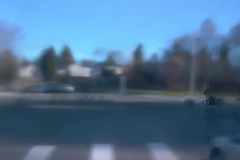} &
\includegraphics[width=\linewidth]{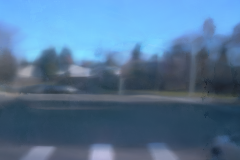} &
\includegraphics[width=\linewidth]{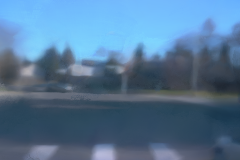} \\
\includegraphics[width=\linewidth]{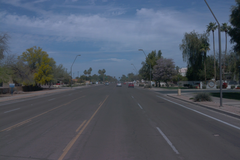} & \includegraphics[width=\linewidth]{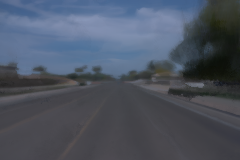} &
\includegraphics[width=\linewidth]{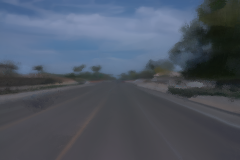} &
\includegraphics[width=\linewidth]{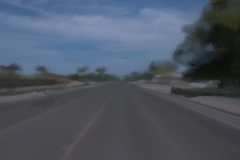} &
\includegraphics[width=\linewidth]{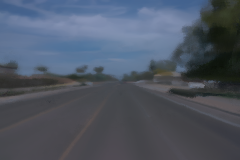} 
\end{tabular}
\caption{Rendered reconstructions of scenes from the Waymo~\cite{waymo} dataset where camera poses are initialized with different levels of noise. Our joint pose refinement step ensures that renderings are clear irrespective of the level of multiplicative noise applied to the camera pose.}\label{fig:noise_recon}

\end{figure*}

\begin{figure*}[t!]
\setlength{\tabcolsep}{2pt}
\newcolumntype{Y}{>{\centering\arraybackslash}p{0.19\textwidth}}
\centering
\begin{tabular}{YYYYY}
Reference & $\mathcal{N}(0,0.1)$ & $\mathcal{N}(0,0.3)$ & $\mathcal{N}(0,0.5)$ & $\mathcal{N}(0,1.0)$ \\
\includegraphics[width=\linewidth]{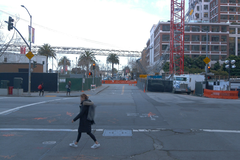} & \includegraphics[width=\linewidth]{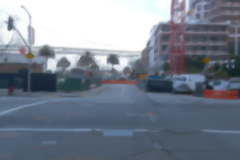} &
\includegraphics[width=\linewidth]{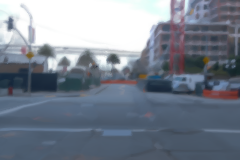} &
\includegraphics[width=\linewidth]{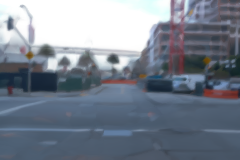} &
\includegraphics[width=\linewidth]{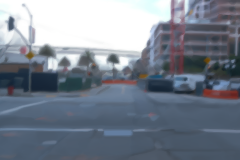} \\
\includegraphics[width=\linewidth]{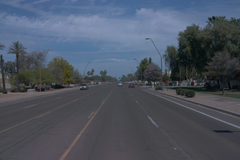} & \includegraphics[width=\linewidth]{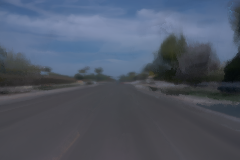} &
\includegraphics[width=\linewidth]{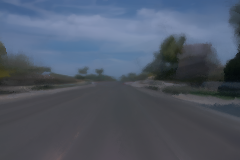} &
\includegraphics[width=\linewidth]{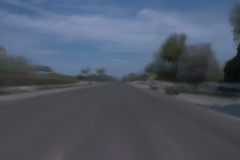} &
\includegraphics[width=\linewidth]{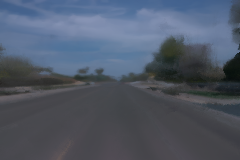} 
\end{tabular}

\caption{Rendered novel views of scenes from the Waymo~\cite{waymo} dataset where camera poses are initialized with different levels of noise. Our method is robust to different levels of multiplicative noise applied to the camera poses describing its suitability for learning scene representations even when the initalized poses are noisy.}\label{fig:noise_novel}

\end{figure*}

To test the robustness of our pose refinement step, we augment the ground truth poses of a subset of our dataset with four different levels of noise, \ie $\delta \sim \mathcal{N}(0, 0.1)$,$\delta \sim \mathcal{N}(0, 0.3)$, $\delta \sim \mathcal{N}(0, 0.5)$, and $\delta \sim \mathcal{N}(0, 1.0)$. We can see in \cref{tab:pose_ablations} and that in each case our refined poses have much lower absolute trajectory errors as compared to the noisy poses. If we compare the renderings for both reconstructions and novel views in \cref{fig:noise_recon} and \cref{fig:noise_novel} respectively, we can see that we get clear renderings irrespective of the amount of noise that was added. This shows that our joint pose refinement is robust to noisy initializations of camera poses.  

\begin{table}[ht!]
  \centering
  \begin{tabular}{@{}c|cccc|cccc|cccc@{}}
    \toprule
     Pose & \multicolumn{12}{c}{ATE (in m) $\downarrow$}\\
     \cmidrule{2-13}
     & \multicolumn{4}{c|}{Mean} & \multicolumn{4}{c|}{Median} &   \multicolumn{4}{c}{Standard Deviation}\\  
 \midrule
    noise level    & $0.1$ & $0.3$ & $0.5$& $1.0$& $0.1$& $0.3$ & $0.5$&$1.0$&$0.1$&$0.3$&$0.5$&$1.0$ \\
    \midrule
    W/o refine & 15.741 & 15.928 & 16.109 & 16.515 & 9.233 & 9.435 & 9.639 & 10.0562 & 15.904 & 16.007& 16.106 & 16.328 \\
    Refined & 0.020 & 0.023 & 0.034 & 0.063 & 0.020 & 0.008 & 0.009 & 0.014 & 0.010 & 0.050 & 0.083 & 0.158\\
    \bottomrule
  \end{tabular}

  \caption{Average absolute trajectory error (ATE) in metres computed before and after pose refinement for different levels of noise. In each case, our joint camera refinement step rectifies inaccuracies in camera pose and achieves much lower trajectory errors as compared to non-refined poses.}
  
  \label{tab:pose_ablations}
  \vspace{-0.4cm}
\end{table}

\end{document}